\documentclass[nohyperref]{article}

\usepackage{microtype}
\usepackage{graphicx}
\usepackage{booktabs} %

\usepackage{hyperref}

\usepackage[accepted]{icml2023}

\usepackage{amsmath}
\usepackage{amssymb}
\usepackage{mathtools}
\usepackage{amsthm}
\usepackage{xspace}

\usepackage{algorithm}
\usepackage{subcaption}
\usepackage{comment}

\usepackage[capitalize,noabbrev]{cleveref}

\theoremstyle{plain}
\newtheorem{theorem}{Theorem}[section]

\theoremstyle{definition}

\theoremstyle{remark}

\newcommand{\x}{\pmb{x}}

\newcommand{\w}{\pmb{w}}
\newcommand{\z}{\pmb{z}}
\newcommand{\xib}{\pmb{\xi}}
\newcommand{\zetab}{\pmb{\zeta}}
\newcommand{\del}{\pmb{\delta}}
\newcommand{\g}{\mathbf{g}}
\newcommand{\HH}{\mathbf{H}}

\newcommand{\LL}{\mathcal{L}}
\newcommand{\F}{\mathcal{F}}
\newcommand{\N}{\mathcal{N}}
\newcommand{\W}{\mathcal{W}}
\newcommand{\M}{\mathcal{M}}
\newcommand{\OO}{\mathcal{O}}
\newcommand{\E}{\mathbb{E}}
\newcommand{\Te}{N}
\DeclareMathOperator*{\argmin}{arg\,min}
\DeclareMathOperator*{\argmax}{arg\,max}

\newcommand{\alg}{\textsc{Crest}\xspace} 
\newcommand{\craig}{\textsc{Craig}\xspace} 
\newcommand{\grad}{\textsc{GradMatch}\xspace} 
\newcommand{\glister}{\textsc{Glister}\xspace}
\newcommand{\ada}{\textsc{AdaCore}\xspace}

\newcommand{\ba}[1]{{{\textcolor{red}{[B: #1]}}}}

\usepackage[textsize=tiny]{todonotes}

\icmltitlerunning{Towards Sustainable Learning: Coresets for Data-efficient Deep Learning}

\begin{document}

\twocolumn[
\icmltitle{Towards Sustainable Learning: Coresets for Data-efficient Deep Learning}

\icmlsetsymbol{equal}{*}

\begin{icmlauthorlist}
\icmlauthor{Yu Yang}{yyy}
\icmlauthor{Hao Kang}{gt}
\icmlauthor{Baharan Mirzasoleiman}{yyy}
\end{icmlauthorlist}

\icmlaffiliation{yyy}{Department of Computer Science, University of California Los Angeles, USA}
\icmlaffiliation{gt}{School of Computer Science, Georgia Institute of Technology}

\icmlcorrespondingauthor{Yu Yang}{yuyang@cs.ucla.edu}
\icmlcorrespondingauthor{Baharan Mirzasoleiman}{baharan@cs.ucla.edu}

\icmlkeywords{Machine Learning, ICML}

\vskip 0.3in
]

\printAffiliationsAndNotice{\icmlEqualContribution} %

\begin{abstract}
To improve the efficiency and sustainability of learning deep models, we propose \alg, the first scalable framework with rigorous theoretical guarantees to identify the most valuable examples for training non-convex models, particularly deep networks. To guarantee convergence to a stationary point of a non-convex function, \alg models the non-convex loss as a series of quadratic functions and extracts a coreset for each quadratic sub-region. In addition, to ensure faster convergence of stochastic gradient methods such as (mini-batch) SGD, \alg iteratively extracts multiple mini-batch coresets from larger random subsets of training data,
to ensure nearly-unbiased gradients with small variances.
Finally, to further improve scalability and efficiency, \alg identifies and excludes the examples that are learned from the coreset selection pipeline. Our extensive experiments on several deep networks trained on vision and NLP datasets, including CIFAR-10, CIFAR-100, TinyImageNet, and SNLI, confirm 
that \alg speeds up training deep networks on very large datasets, by 1.7x to 2.5x with minimum loss in the performance.
By analyzing the learning difficulty of the subsets selected by \alg, we show that deep models benefit the most by learning from subsets of increasing difficulty levels
\footnote{Code can be found at \href{https://github.com/bigml-cs-ucla/crest}{https://github.com/bigml-cs-ucla/crest}}.\looseness=-1
\end{abstract}

\section{Introduction}
Large datasets have enabled over-parameterized neural networks to achieve unprecedented success \cite{devlin2018bert, brown2020language, zhai2022scaling}. %
However, training such models, with millions or billions of parameters, on large data %
requires expensive computational resources, which consume substantial energy, leave a massive amount of carbon footprint, and often soon become obsolete and turn into e-waste \citep{asi2019importance,schwartz2019green,strubell2019energy}.
While there has been a persistent effort to improve the performance and reliability of machine learning models \citep{brown2020language,xiao2015learning,zhai2022scaling}, their sustainability is often neglected. %

Indeed, not all examples are equally valuable or even required to guarantee a good generalization performance.
To address the sustainability and efficiency of machine learning, one approach involves selecting the most relevant data for training. 
A recent line of work \cite{mirzasoleiman2020coresets,killamsetty2021glister,killamsetty2021grad,pooladzandi2022adaptive} showed that for strongly convex models, a weighted subset (coreset) of data that closely matches full gradient---sum of the gradient of all the training examples---provide convergence guarantees for (Incremental) Gradient Descent. %
Such coresets speed up learning by up to 6x.
Intuitively, this is possible as for popular strongly convex models|logistic, linear regression, and regularized support vector machines|%
the gradient error of a coreset
during the \textit{entire training} %
can be upper-bounded in advanced \cite{mirzasoleiman2020coresets}.

Unfortunately, we cannot simply apply the same idea to non-convex models, for three reasons.
First, for non-convex models, training dynamics|loss and gradient of examples|drastically change during the training and cannot be upper-bounded beforehand. 
As a result, the importance of examples for learning changes throughout training, and one coreset cannot guarantee convergence anymore.
Second, non-convex models are learned with (mini-batch) stochastic gradient methods, such as  (mini-batch) SGD, which require unbiased estimates of the full gradient with a bounded variance. Existing coresets %
that capture the gradient of the full data cannot provide any guarantee for stochastic gradient methods, as the gradient of mini-batches selected from the coreset may be biased and have a large variance. %
Finally, iteratively selecting coresets from full data %
becomes very expensive for large datasets, and does not yield speedup.

In this work, we address the above challenges and propose \alg, a rigorous method to find coresets for %
non-convex models, by making the following contributions:

\textbf{Coreset selection by modeling the non-convex loss.} %
To ensure a small gradient error throughout training, our key idea is to divide the loss into multiple quadratic sub-regions and find a coreset for learning every quadratic sub-region.
To do so, we model the loss of every example 
as a quadratic function based on its current gradient and curvature information at model parameters $\w_{t_l}$. 
Then, we find a coreset that captures the full gradient %
at $\w_{t_l}$, and keep training on it as long as the quadratic approximated loss of the coreset (sum of quadratic functions corresponding to its elements) closely captures the actual loss of %
the full data. 
Otherwise, we update the coreset. In doing so, we ensure a small gradient error during the entire training, which we leverage to guarantee convergence to a stationary point.
    
\textbf{Coresets for (mini-batch) stochastic gradient methods.} To address coreset selection for (mini-batch) stochastic gradient methods, our idea is to sample multiple subsets of training data uniformly at random, and select a mini-batch coreset from every random sample to closely capture its gradient. 
The gradients of larger random subsets are unbiased estimates of the full gradient, but have a considerably smaller variance.
Hence, the mini-batch coreset gradients are nearly unbiased, and have a small variance. 
Updating the mini-batch coresets based on above piece-wise quadratic loss approximation, ensures a small bias throughout training.
This allows providing superior convergence guarantee for training with stochastic gradient methods. Besides, it significantly improves the computational complexity of finding coresets and scales coreset selection to much larger datasets. \looseness=-1

\textbf{Further improving the efficiency of coreset selection.}
To further improve the efficiency and scalability of coreset selection, we make the following observation. 
When a group of examples are learned, their gradients become nearly zero. Hence, a few examples can well represent the gradient of the corresponding group and the entire group can be safely excluded from coreset selection afterwards. 
\alg iteratively excludes examples that are learned and have a very small loss during multiple consecutive training iterations, and finds mini-batch coresets from the remaining examples.
This speeds up learning, and %
improves the  efficiency and performance of coreset selection, in later stages of training. %

Through extensive experiments, we demonstrate the effectiveness of \alg for training various over-parameterized models on different vision and NLP benchmark datasets, including ResNet20 on CIFAR-10 \cite{cifar10}, ResNet18 on CIFAR-100 \cite{cifar10}, ResNet-50 on TinyImageNet \cite{imagenet15russakovsky}, and RoBERTa \cite{liu2019roberta} on SNLI \cite{bowman-etal-2015-large} with 570K examples.
\alg is able to achieve 1.7x to 2.5x speedup over training on the full data, while introducing the smallest relative error compared to the baselines.
To our knowledge, this is the first time coreset selection has been applied to such large models and datasets in vision and NLP.

Finally, we analyze the examples selected by \alg at different times during the training. We quantify the learning difficulty of every example using the forgettability score 
\cite{toneva2018empirical}, which counts the number of times an example is misclassified after being correctly classified during training.
We find that early in training, the most effective subsets for learning deep models are easy-to-learn examples. As training proceeds, the model learns the most from examples with increasing levels of learning difficulty. Interestingly, the model never requires training on easiest-to-learn examples to achieve a good generalization performance.
\section{Related Work}
Several heuristics have been recently proposed for finding coresets for training machine learning models. 
A line of work first fully trains the original model \cite{birodkar2019semantic} or a smaller proxy \cite{coleman2020selection}. Then, it selects examples with the most centrally located embeddings  \cite{birodkar2019semantic}, highest uncertainty, i.e., the entropy of predicted class probabilities \citep{coleman2020selection}, %
largest forgetting events, i.e., the number of times an example is misclassified after being correctly classified \citep{toneva2018empirical}, or large expected gradient norm over multiple initializations \cite{paul2021deep}.
These methods do not yield any speedup or theoretical guarantees.

Another line of work selects examples during training to speed up learning.
Importance sampling techniques employ the gradient norm \citep{alain2015variance,katharopoulos2018not} or the loss \citep{ loshchilov2015online,schaul2015prioritized} to %
reduce the variance of %
stochastic gradients during the training. However, importance sampling does not provide rigorous convergence guarantees and cannot provide a notable speedup for training deep models. 
\citet{mindermann2022prioritized} finds examples that are non-noisy, non-redundant, task-relevant, and reduce the loss on a holdout set the most. This method speeds up training but requires a validation set and does not guarantee convergence.

Most relevant to our work are recent theoretically rigorous techniques that select coresets by iteratively matching the (preconditioned) gradient of full training data, namely \cite{mirzasoleiman2020coresets,killamsetty2021grad,pooladzandi2022adaptive}, or validation set \cite{killamsetty2021glister}.
Such methods guarantee convergence to a near-optimal solution, for training (strongly) convex or nearly convex models under Polyak-Lojasiewicz (PL) condition, using Incremental Gradient (IG) methods, or Gradient Descent (GD). %
However, they do not guarantee convergence for training non-convex models trained with (mini-batch) stochastic gradient methods, and do not scale to very large datasets.
Our work addresses the above shortcomings by developing a rigorous and scalable framework to extract coresets for data-efficient deep learning.\looseness=-1
\section{Problem Formulation and Background
}
The standard approach to training machine learning models is empirical risk minimization (ERM). Formally, given a loss function $\LL$, we find the model parameters $\w$ that minimize the expected loss on training examples $\{(\x_i,y_i)\}_{i=1}^n$ indexed by $V \!=\! \{1, \cdots\!, n\}$, sampled from distribution $\mathcal{D}$: 
\begin{equation}
    \w^*\!=\!\argmin_{\w\in\W} \LL(\w)\!:= \mathbb{E}_{(\x_i,y_i)\sim \mathcal{D}}[\LL(\w; (\x_i, y_i))].\vspace{-2mm}
\end{equation}
For over-parameterized models trained on large training data, GD becomes prohibitively slow. Hence, stochastic gradient methods, such as mini-batch SGD are employed in practice. Such methods select one or a mini-batch $\M$ of $m$ examples sampled i.i.d. from the  training data, and iteratively step in the negative direction of the stochastic gradient of the sampled examples, scaled by step-size $\eta$:
\begin{equation}\vspace{-2mm}
    \w_{t+1}=\w_t-\eta \frac{1}{m}\sum_{i\in\M}\g_{t,i},\vspace{-2mm}
\end{equation}
where $\g_{t,i}\!=\!\nabla \LL_i(\w_t)\!:=\!\nabla \LL(\w_t;(\x_i,y_i))$ is the gradient of example $i$ at iteration $t$. 
Random examples have an unbiased gradient with a bounded variance, i.e., $\E_{i\in V}[\|\g_{t,i}-\nabla \LL(\w_t)\|^2]\leq \sigma^2$. Hence, they guarantee convergence with an $\OO(1/\sqrt{t})$ rate to a stationary point of a non-convex loss \cite{ghadimi2013stochastic}.
Importantly, random mini-batches of size $m$ have an unbiased gradient with a reduced variance of $\sigma^2/m$. As long as mini-batches are not too large, mini-batch SGD achieves a faster convergence rate of $\OO(1/\sqrt{mt})$ \cite{wang2019stochastic,jin2021nonconvex}.

Existing coreset methods, such as \craig \cite{mirzasoleiman2020coresets}, %
\grad \cite{killamsetty2021grad}, and \ada \cite{pooladzandi2022adaptive} find weighted subsets of examples that match the full training gradient (preconditioned on Hessian). Formally, the goal is to find the smallest subset $S \subseteq V$
and corresponding per-element step-sizes (weights) $\gamma_j > 0$ that approximate the full gradient with an error at most $\epsilon > 0$ for all the possible values of %
$\w_t \in \W$:
\begin{align}\label{eq:error}\vspace{-2mm}
\hspace{-3mm}S^*\!\!=\!\!\hspace{-4mm}\argmin_{S \subseteq V, \gamma_j \geq 0~ \forall j\in S} \hspace{-4mm}|S|, 
~\text{s.t.}~   
\max_{\w_t\in\W}\! \|\! \sum_{i\in V} \g_{t,i} \!-\!\! \sum_{j \in S} \gamma_{j} \g_{t,j} \| \!\leq\! \epsilon. \vspace{-2mm}
\end{align}
Problem (\ref{eq:error}) requires calculating the maximum gradient error between full and coreset gradient for all $\w_t\in \W$, which cannot be computed. To address this, \citet{mirzasoleiman2020coresets} showed that for several classes of (strongly) convex problems, including regularized linear and ridge regression, and support vector machines (SVMs), the normed gradient difference between data points \textit{during the entire training} can be efficiently %
upper-bounded by
the difference between feature vectors. %
This allows turning Problem \eqref{eq:error} into the following submodular\footnote{A set function $F:2^V \rightarrow \mathbb{R}^+$ is submodular if $F(S\cup\{e\}) - F(S) \geq F(T\cup\{e\}) - F(T),$ for any $S\subseteq T \subseteq V$ and $e\in V\setminus T$.  
$F$ is \textit{monotone} if $F(e|S)\geq 0$ for any $e \! \in \! V \! \setminus \! S$ and $S \subseteq V$.} \!cover problem: %
\begin{equation}\label{eq:cover}
    S^*\!\!=\!\argmin|S| ~~~ \text{s.t.} ~~~ C-\!\sum_{i\in V}\min_{j\in S}\|\x_i-\x_j\|\geq C-\epsilon,
\end{equation}
where $C$ is a big constant. A near-optimal coreset of size $k$ can be found from a ground-set of $n$ elements, using the greedy algorithm with complexity of $\mathcal{O}(n\!\cdot\! k)$ as a pre-processing step before training. %
The weights $\gamma_j$ are calculated as the number of examples $i\!\in\! V$ for which $j\!\in\! S$ minimizes $\|\x_i-\x_j\|$.
This approach has been adopted by \cite{killamsetty2021glister, pooladzandi2022adaptive}.
\citet{killamsetty2021grad} used orthogonal matching pursuit (OMP) to directly find a weighted coreset, by minimizing the regularized objective in RHS of Problem (\ref{eq:error}). However, OMP provides weaker guarantees than greedy, and does not always find a large enough subset. Hence, the coreset needs to be augmented with random examples. %

For neural networks, %
finding coresets based on their very high-dimensional gradients is slow and does not yield high-quality coresets.
Instead, one can use the gradient of the loss w.r.t the input to the last layer that is shown to capture the variation of the gradient norm well \cite{katharopoulos2018not}.
Such lower-dimensional %
gradients $\g^L_{t,i}$ can be  quickly obtained with a forward pass and can be used instead of the full gradient to find coresets during the training %
\cite{mirzasoleiman2013distributed,pooladzandi2022adaptive,killamsetty2021glister}. 
Moreover, with a fixed training budget one can find a subset of size $k$ by solving the following submodular maximization problems, which is the dual of the submodular cover Problem \eqref{eq:cover}: %
\begin{equation}\label{eq:sub_max}
    S_{t}^*\!=\argmax_{S\subseteq V} \!~C\!-\!\!\sum_{i\in V}\min_{j\in S}\| \g^L_{t,i}\!-\g^L_{t,j} \| , ~ \text{s.t.} ~ |S| \!\leq\! k.
\end{equation}
However, it is not clear when the coresets should be updated to
guarantee convergence for training non-convex models. Besides, finding coresets from the full data does not scale to training on large datasets. Importantly, the above method only guarantees convergence for (Incremental) GD and cannot guarantee convergence for stochastic gradient methods used for training neural networks, as we will discuss next.\looseness=-1

\section{Coresets for Training Non-convex Models}
\begin{figure*}[t]
     \centering
      \begin{subfigure}[t]{0.24\textwidth}
         \centering
         \includegraphics[ width=\textwidth]{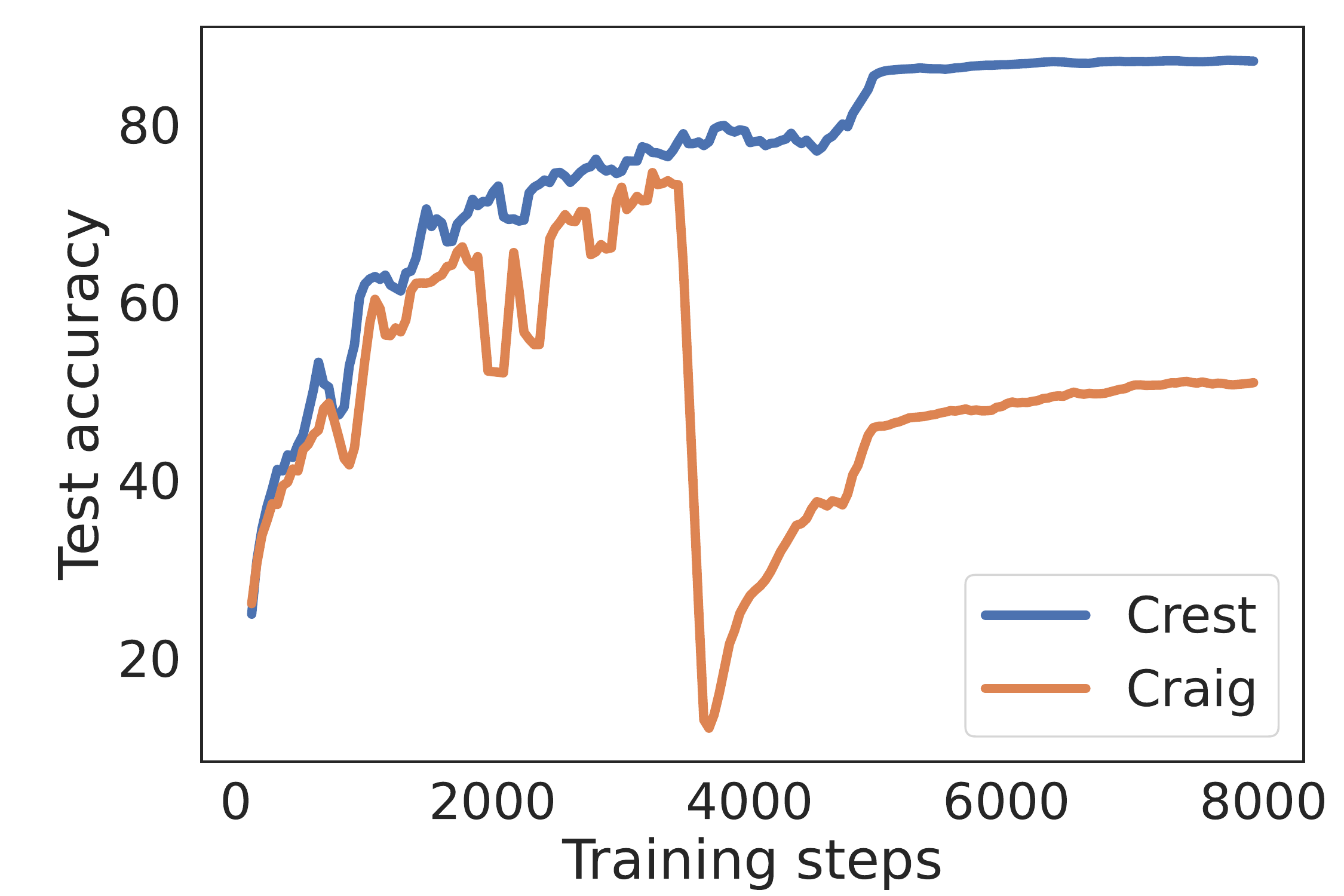}\vspace{-1mm}
         \caption{Test Accuracy}
         \label{fig:fail_acc}
     \end{subfigure}
     \begin{subfigure}[t]{0.24\textwidth}
         \centering
         \includegraphics[width=\textwidth]{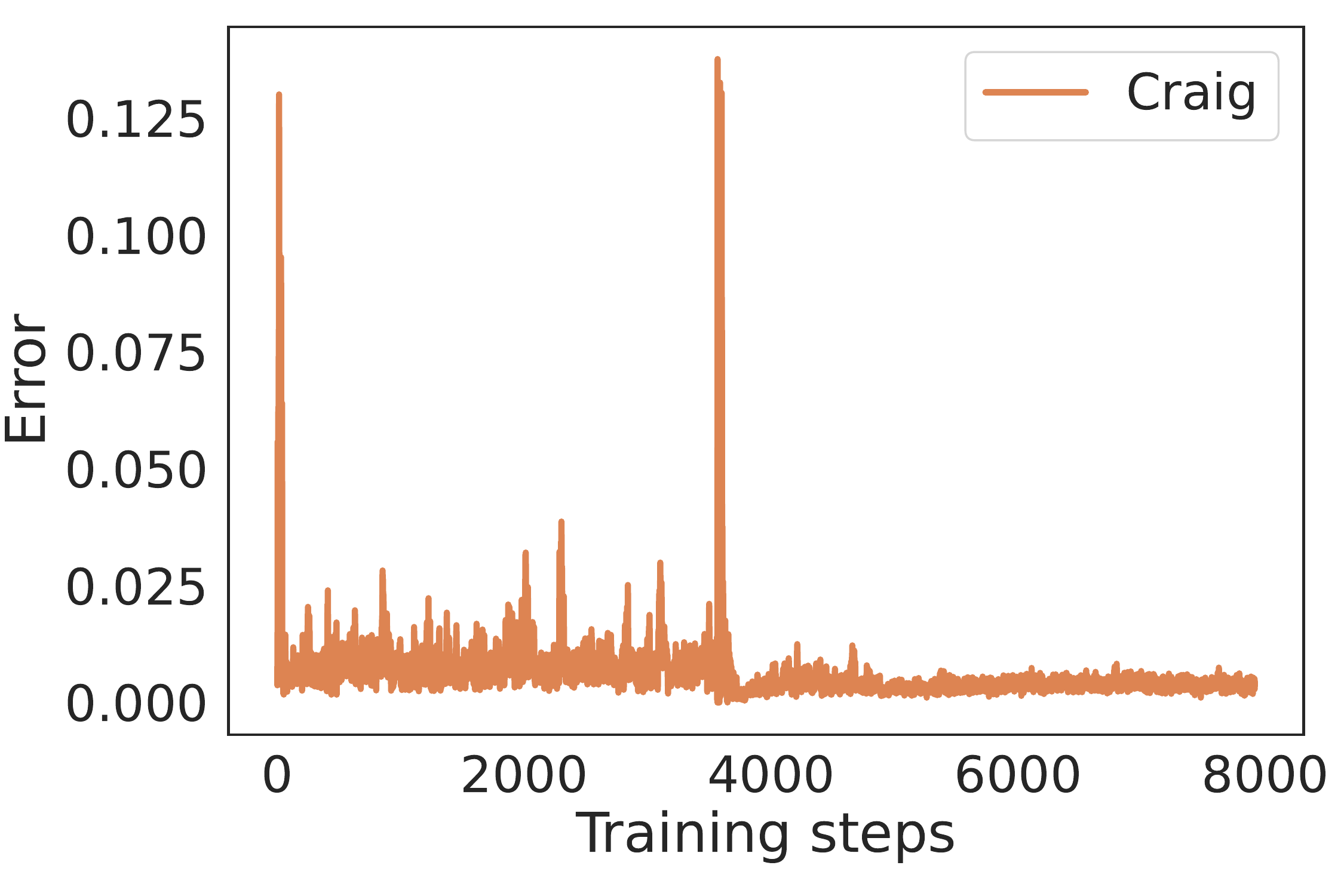}\vspace{-1mm}
         \caption{Error of coreset gradients 
        }
         \label{fig:error}
     \end{subfigure}
     \begin{subfigure}[t]{0.24\textwidth}
         \centering
         \includegraphics[width=\textwidth]{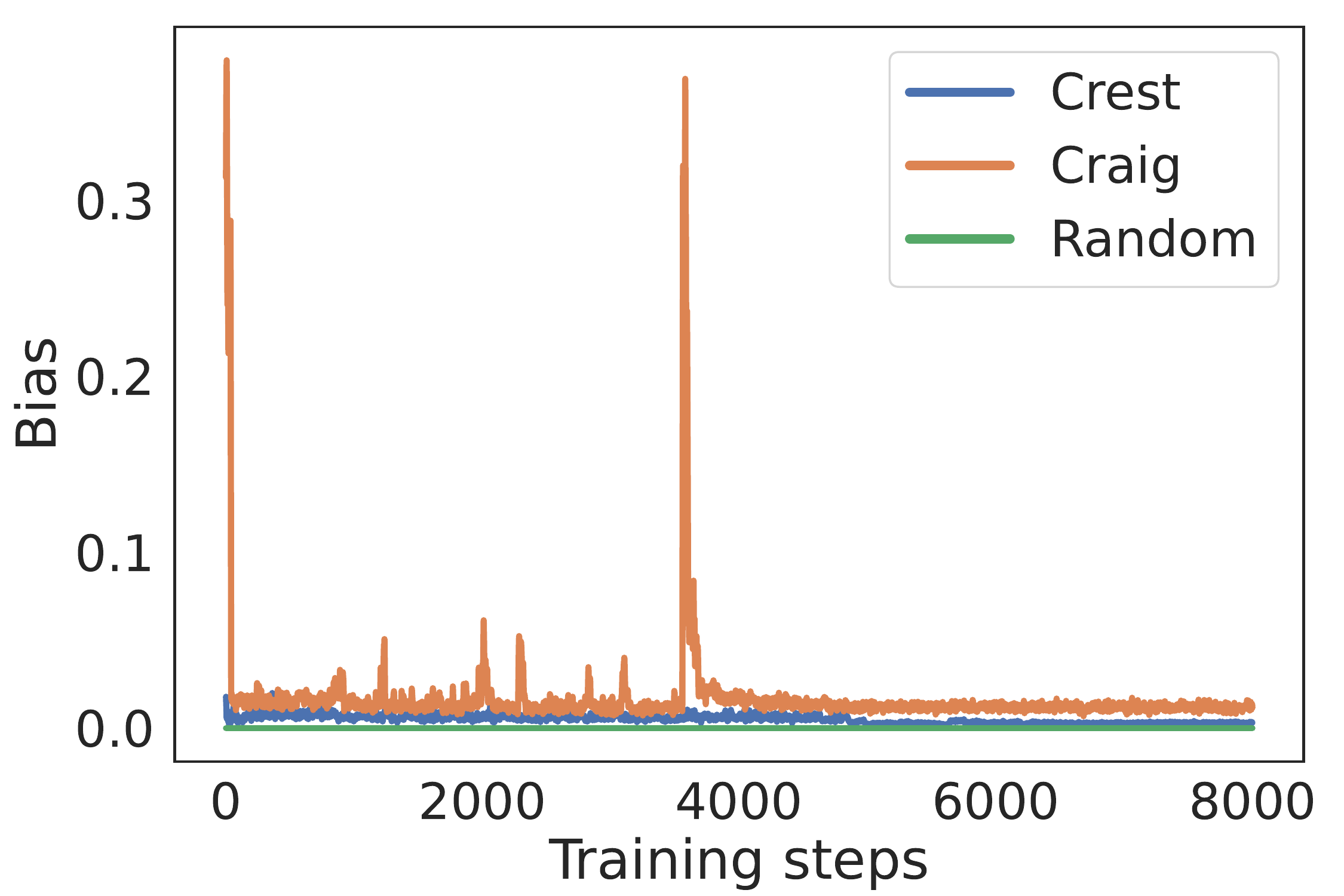}\vspace{-1mm}
         \caption{Bias of mini-batch grads. 
        }
         \label{fig:1}
     \end{subfigure}
     \begin{subfigure}[t]{0.24\textwidth}
         \centering
         \includegraphics[width=\textwidth]{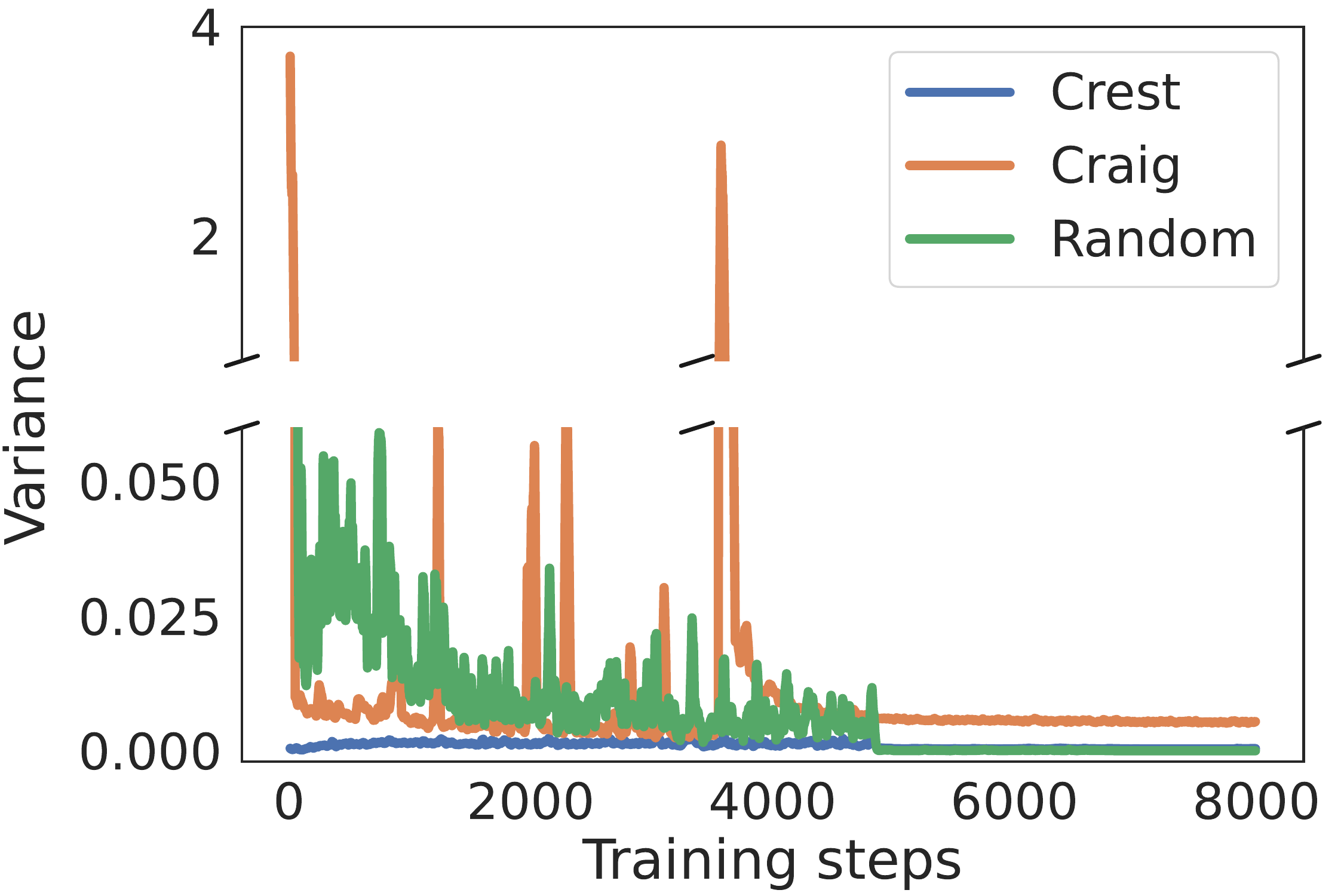}\vspace{-1mm}
         \caption{Variance of mini-batch grads.
        }
         \label{fig:2}
     \end{subfigure}
      \vspace{-2mm}
        \caption{Training ResNet20 on CIFAR-10. (a) 10\% \craig coresets selected at the beginning of every epoch from full data may perform very poorly. This is because,  \textbf{(C1)}: (b) Coresets may have a large error: $\|\g_{t,S}\!-\!\nabla \LL(\w_t)\|$, after a few training iterations; and
        \textbf{(C2)}: Gradient of weighted mini-batches selected from the coresets may have a (c) large bias $\|\E_i[\g_{t,\M_i}]\!-\!\nabla \LL(\w_t)\|$ and (d) large variance $\E_i[\|\g_{t,\M_i}\!-\!\nabla \LL(\w_t)\|^2]$, where $\M_i\in S$ is a mini-batch and $\g_{t,\M_i}\!=\!\E_{j\in \M_i}[\gamma_j \g_{t,j}]$. In contrast, our \alg coresets are nearly unbiased, and have a smaller variance than random mini-batches of same size. \looseness=-1
        }
        \label{fig:conv_core}
        \vspace{-4mm}
\end{figure*}
In this section, we will first discuss the challenges of extracting coresets for deep models, and then introduce our proposed method, \alg, to overcome the above challenges and make coreset selection applicable to neural networks.

\vspace{-3mm}\paragraph{Challenges.}
The %
non-convex loss landscape of the non-convex models makes coreset selection very challenging. {Fig. \ref{fig:fail_acc}} shows that %
existing coreset selection methods such as \craig \cite{mirzasoleiman2020coresets} %
that find coresets by iteratively solving Eq. \eqref{eq:sub_max} at %
every epoch
may perform very poorly for training deep networks,
for the following reasons: \looseness=-1

\textbf{(C1)} For deep networks, the loss functions associated with different data points $\LL_i$ change very rapidly \cite{defazio2019ineffectiveness}.
Therefore, in contrast to (strongly) convex functions, for which the %
gradient error of a coreset
throughout training can be effectively upper-bounded in advance, e.g., using their feature vectors, such upper bounds cannot be computed for neural networks. %
That is, even within a relatively small neighborhood $\N$ around $\w_t$, %
the gradient $\nabla \LL_i(\w_t)$ may be drastically different than $\nabla \LL_i(\w_t\!+\!\del)$ for $\w_t\!+\del\!\in\! \N$. 
\cref{fig:error} shows that the gradient error of coresets found by \craig can be very large after a few training iterations.
Here, the challenge is to compute the size of the neighborhood in which a coreset closely captures the full gradient, and update the coreset otherwise.

\textbf{(C2)} Coresets found from the full training data 
guarantee convergence for (Incremental) GD, but
cannot provide any guarantee for \textit{stochastic} gradient methods, such as (mini-batch) SGD, that are applied to train neural networks. This is because 
stochastic methods require {unbiased} estimates of the full gradient with a bounded variance. %
However, as the error of the coresets found from the full data may increase during the training, the gradient of mini-batches selected from the coresets may have a large bias.
Besides, %
as some examples may have a very large weight, 
the variance of weighted mini-batch gradients %
are much larger than the variance of random (unweighted) mini-batch gradients used when training on the full data. 
\cref{fig:1}, \ref{fig:2} show that %
gradient of
mini-batches
selected from the coreset can have a very large bias and variance.
Here, the challenge is to find coresets that are nearly unbiased and have a small variance.

\textbf{(C3)} For deep networks, the importance of examples for learning changes over time and hence the coresets should be updated frequently. The greedy algorithm has a complexity of $\OO(n\cdot k)$ to find $k$ out of $n$ examples. For large datasets, this prevents the coreset selection methods to achieve a significant speedup. Hence, the challenge is to improve the efficiency of coreset selection for training deep networks. %

Next, we discuss how we overcome %
the above challenges.\looseness=-1

\subsection{Modeling the Non-convex Loss Function}\label{sec:model}
To address the challenge \textbf{(C1)} of finding the size of the neighborhood in which a coreset closely captures the full gradient,
we model the non-convex loss %
as a piece-wise quadratic function.
In doing so, we reduce the problem of finding coresets for a non-convex objective to finding coresets for a series of quadratic problems. 
Formally, at every coreset selection step $l$, we find a coreset $S_l$ that captures the full gradient 
at $\w_{t_l}$. 
Then, we make a quadratic loss approximation $\F^l$ based on the gradient and curvature of the coreset at $\w_{t_l}$. We keep training on the coreset $S_l$ within the neighborhood $\N_l$ in which the quadratic approximation closely follows the actual training loss,
i.e., we have that $ \LL(\w_{t_l}\!+\del) \!=\! \F^l(\del)~\forall ~\w_{t_l}\!+\del\!\in\!\N_l$. %
Otherwise, we update the coreset and make a new quadratic approximation.
This ensures a small gradient error within $\N_l$, and similarly through the entire training. Hence, convergence can be guaranteed.

In this work, we extract the coresets by greedily solving the submodular Problem \eqref{eq:sub_max}.
But, our piece-wise quadratic approximation can be generally applied to any coreset selection method to check the validity of the coresets, and updating them to guarantee convergence for deep learning. \looseness=-1

In the rest of this section, we first discuss how to efficiently estimate %
the coreset loss %
as a quadratic function $\F^l$, %
based on its
gradient and curvature at $\w_{t_l}$.
Then, we discuss finding the size of the neighborhood $\N_l$ in which the quadratic function $\F^l$ closely captures the loss $\LL$ of full  training data.

\vspace{-2mm}
\paragraph{Approximating coreset losses by quadratic functions.} 
We model the %
coreset loss within the neighborhood $\N_l$ by a quadratic approximation $\F^l$, using the $2^\text{nd}$-order Taylor series of expansion of $\LL(\w_{t_l})$ %
at $\w_{t_l}$ where the coreset is extracted:
\vspace{-2mm}
\begin{align}\label{eq:m}
    \F^{l}(\del)=\frac{1}{2}\del^T\HH_{t_l,S_l}\del+\g_{t_l,S_l}\del
    +\LL(\w_{t_l}), %
\end{align}
where $\HH_{t_l,S_l}\!\!=\!\!\frac{1}{|S_l|}\!\sum_{j\in S_l}\gamma_j\HH_{t_l,j}$ and $\g_{t_l,S_l}\!\!=\!\!\frac{1}{|S_l|}\allowbreak\sum_{j\in S_l}\gamma_j \g_{t_l,j}$ are the weighted mean of the Hessian and gradient of the examples in the coreset $S_l$.
Such modeling 
is the main idea behind the popular %
convexification technique in mathematical optimization, which powers %
Levenberg-Marquardt \cite{marquardt1963algorithm} and K-FAC \citep{martens2015optimizing} optimization methods, among others \cite{bertsekas1979convexification,carmon2018accelerated,wang2019stochastic}. \looseness=-1

To obtain an efficient estimate of the Hessian of the coreset, %
we use an approximate Hessian operator instead of the full Hessian.
Specifically, we employ the Hutchinson's trace estimator method \cite{hutchinson1989stochastic} to obtain a stochastic estimate of the coreset Hessian diagonal \cite{bollapragada2019exact,dembo1982inexact,xu2020newton,yao2018inexact}, without having to form the Hessian matrix explicitly: 
\begin{align}\label{eq:hf}
\text{diag}(\HH_{t_l,S_l})=\mathbb{E}[\z\odot (\HH_{t_l,S_l}\z)]. 
\end{align}
This method approximates the Hessian diagonal as the expectation of Hessian $\HH_{t_l,S_l}$ multiplied by a random vector $\z$  with Rademacher distribution. The multiplication $\HH_{t_l,S_l} \z$ can be efficiently calculated via backprop on gradients of the coreset multiplied by $\z$. i.e., $\HH_{t_l,S_l}\z=\partial \g_{t_l,S_l}^T \z /\partial \w_{t_l}$.

As the local gradient and curvature information can be very noisy for neural networks \cite{yao2020adahessian}, %
to better approximate the global gradient and Hessian information, we smooth them out
by applying exponential averaging with parameters $0<\beta_1$, $0<\beta_2<1$:  %
\begin{align}\label{eq:g_avg}
\overline{\g}_{t_l,j}&=\frac{(1-\beta_1)\sum_{t=1}^{t_l} \beta_1^{t_l-t}\g_{t,j}}{1-\beta_1^{t_l}}, 
\end{align}\vspace{-4mm}
\begin{align}\label{eq:h_avg}
\hspace{-2mm}\overline{\HH}_{j,t_l}\!=\!\sqrt{\!\frac{(1\!-\!\beta_2)\sum_{t=1}^{t_l} \beta_2^{t_l-t} \text{diag}(\HH_{t,j})\text{diag}(\HH_{t,j})}{1-\beta_2^{t_l}}}.
\end{align}
For very large networks, %
gradient and Hessian diagonal w.r.t. the input to the penultimate layer can be used in Eq. (\ref{eq:hf}-\ref{eq:h_avg}).\looseness=-1 %

\paragraph{Estimating the size of the quadratic neighborhoods.}
To check the validity of the coreset, we iteratively compare the value of the quadratic loss %
$\F^{l}(\del)$ with the value of the actual training loss. For efficiency, we obtain an unbiased estimate of the actual loss on a small random sample of training examples $V_r\!\subseteq \!V$, i.e., $\LL^r$. 
We update the coreset $S_l$ and the quadratic approximation $\F^l(\del)$, when the quadratic coreset loss does not closely follow the actual loss estimate $\LL^r(\del+\w_{t_l})$.
More precisely, every $T_1$ iterations, we compute the ratio of the absolute loss difference to the actual loss, i.e.,  %
\begin{equation}\label{eq:threshold}
\rho_{t_l}\!=\!\frac{|\F^l(\del)-\LL^r(\del\!+\!\w_{t_l})|}{\LL^r(\del+\w_{t_l})}.
\end{equation}
We consider the quadratic approximation of the coreset loss %
to be sufficiently accurate if $\rho_{t_l}$ is smaller than a threshold $\tau$. If $\rho_{t_l}\leq\tau$, we keep using the coreset for the $T_1$ subsequent iterations. 
Otherwise, we find a new coreset and update the quadratic approximation, accordingly.
Computing $\rho_{t_l}$ can be done quite efficiently. $\F^l(\del)$ %
can be efficiently calculated based on the gradient and Hessian of the coreset using Eq. (\ref{eq:hf}-\ref{eq:h_avg}). 
$\del$ %
is the total amount of updates calculated by the optimization algorithm in $T_1$ training iterations.
Calculating $\LL^r(\del\!+\!\w_{t_l})$ requires an additional forward pass {on a subset $V_r$ of data}, which we only need once every $T_1$ iterations. \looseness=-1 

\textbf{Remark.} In the {initial} phase of training,  gradients evolve very rapidly.
Hence, %
early in training, the quadratic approximations are accurate in a small neighborhood $\N_l$. 
Therefore, it is crucial to update the coresets frequently to be able to closely capture the full gradient. %
In contrast, in the final stage of training, the loss becomes well approximated as a convex quadratic within a sufficiently large neighborhood of the local optimum \cite{martens2015optimizing}. 
Hence, the same subset can be used for several training iterations. 
We show in our experiments that for a fixed $\tau$, \alg updates the coresets much less frequently as training proceeds.
In practice, $T_1$ can grow proportional to the inverse of the norm of the Hessian diagonal, as we confirm experimentally.

\subsection{Coresets for (Mini-batch) Stochastic GD}
\label{sec:subset}

Next, we address the challenge \textbf{(C2)} of finding coresets for (mini-batch) stochastic gradient methods, that are used for training deep networks.
To address this problem, our main idea is 
to sample multiple subsets of training data $\{V_1,\cdots,V_P\}$ uniformly at random, and directly select a smaller coreset $S_l^p, p\in[P]$ of the mini-batch size $m$ %
from each random subset $V_p$.
Effectively, instead of selecting a subset to capture the full gradient at $\w_{t_l}$, we select {multiple} \textit{mini-batch coresets} $\{S_l^1,\cdots,S_l^P\}$ %
at $\w_{t_l}$, where each coreset $S_l^p$ is of mini-batch size $m$, and captures the full gradient of a random subset $V_p$ of training data at $\w_{t_l}$.

Formally, at every coreset selection iteration $l$, we solve $P$ smaller submodular maximization problems. I.e. $\!$for $p\!\in\![P]$:\looseness=-1 %
\begin{align}\label{eq:mini}
    \hspace{-2mm}{S_{l}^{p}}^*\!\!\!=\!\argmax_{S\subseteq V_p} \!~C\!-\!\!\!\sum_{i\in V_p}\!\!\min_{j\in S}\| \g^L_{t_l,i}\!\!-\!\g^L_{t_l,j} \| , ~ \text{s.t.} ~ |S| \!\leq\! m,
\end{align}
where $\g^L_{t_l,i}$ is the gradient of the loss w.r.t. the input to the last layer of the network at $\w_{t_l}$. 

Then, we make a quadratic loss approximation of the form Eq. \eqref{eq:m} to the \textit{union} of mini-batch coresets $S_l=\bigcup_{p \in [P]} S_l^p$.
Each random subset $V_p$ provides an unbiased estimate of the full gradient, and since each mini-batch coreset $S_l^p$ closely captures the gradient of $V_p$, it provides a nearly unbiased estimate of the full gradient. Therefore, the union of the mini-batch coresets $S_l$ also captures the full gradient. 
However, $S_l$ has a smaller error in capturing the full gradient compared to each of the mini-batch coresets, as small errors of mini-batch coresets cancel each other out 
(\textit{c.f.} \cref{fig:union} in Appendix \ref{appendix:details}). %
Hence, the union of mini-batch coresets makes a more accurate approximation to the {full} loss. 

As long as the quadratic approximation is valid, we can train on \textit{any} of the mini-batch coresets found at $w_{t_l}$.
Thus, we keep %
selecting mini-batch coresets at random from $\{S_l^1,\cdots\!,S_l^P\!\}$, and training on them, as long as the quadratic approximation on the {union} of selected mini-batch coresets %
accurately captures the full loss %
according to Eq. \eqref{eq:threshold}. \looseness=-1

Notably, mini-batch coresets selected from random subsets are nearly unbiased and have a very small variance (\textit{c.f.} \cref{fig:1}, \ref{fig:2}). 
This is because random subsets $V_p$ of size $r$ are unbiased and have a $r/m$ times smaller variance than that of random mini-batches of size $m$. %
As long as %
random subsets are not too large,
mini-batch coresets capture the gradient of random subsets very closely. This ensures that the gradients of mini-batch coresets are nearly unbiased and have a nearly $r/m$ times smaller variance than random mini-batches of same size (\textit{c.f.} %
\cref{fig:3_full} in Appendix).
Note that there is a trade-off. 
For a fixed mini-batch size, selecting mini-batch coresets from larger random subsets results in a smaller variance but may introduce a larger bias.
\begin{algorithm}[tb]
   \caption{CoREsets for STochastic GD (\alg)}
   \label{alg}
\begin{algorithmic}
   \REQUIRE Model parameter $\w_0$, %
   mini-batch size $m$,
   random partition size $r$,
   learning rate $\eta$, total training iterations $\Te$, checking interval $T_2$, multipliers $b,h$, thresholds $\alpha, \tau$. 
   \STATE
   $t \leftarrow 0$, $T_1 \leftarrow 1$, update $\leftarrow$ 1
   \WHILE{$t < \Te$ }
    \IF{update $== 1$}
   \FOR{$p=1$ to $P$}
        \STATE \hspace{0mm}Select a random subset $V_p \subseteq V$ s.t. $|V_p|= r$ 
        \STATE \hspace{0mm}${S_l^p}\!\!\in\!\!\argmax_{\substack{S\subseteq V_p\\ |S|\leq m}}\!\! C\!-\!\!\sum_{i\in V_p}\!\!\min_{j\in S} \!\|\g_{t_l,i}^L\!-\g_{t_l,j}^L\|$ 
   \ENDFOR
       \STATE $S_l=\bigcup_{p \in [P]} S_l^p$ 
       \STATE Calculate $\F^l$ with $\overline{\mathbf{H}}_{t,S_l}, \overline{\g}_{t,S_l}$  %
   \ENDIF
   \FOR{$j = 1$ to $T_1$}
        \STATE $\w_{t+1} \leftarrow \w_t-\eta \g_{S_l,t}$ 
        \STATE $t\leftarrow t+1$
       \IF{$t \mod T_2 == 0$}
       \STATE $\!V\!=\!\{j\!\in\! V | \LL_j(\w_i) \!>\! \alpha, \forall i\in[t\!-\!T_2,t]\}$. 
       \ENDIF
    \ENDFOR
    \STATE $\del\leftarrow\w_{t}-\w_{t-T_1}$
    \STATE Calculate $\rho_t$ from \cref{eq:threshold}.
    \IF{$\rho_t > \tau$} 
       \STATE update $\leftarrow$ 1, 
       \STATE $T_1 \leftarrow h \times \|\overline{\mathbf{H}}_{0}\|/\|\overline{\mathbf{H}}_{t}\|$,~ 
       $P\leftarrow b \times T_1$ 
    \ELSE 
        \STATE update $\leftarrow$ 0
   \ENDIF
   \ENDWHILE
\end{algorithmic}
\end{algorithm}
The very small bias of \alg mini-batch coresets allows guaranteed convergence to a stationary point.
At the same time, their smaller variance ensures superior convergence rate compared to training on full data, %
as we will show in Theorem \ref{thm}.
This cannot be achieved by coresets capturing the full gradient. %

Note that selecting mini-batches from smaller random partitions speeds up the coreset selection, by breaking one large problem into smaller ones. For example, using the greedy algorithm to solve the submodular maximization Problem (\ref{eq:sub_max}) has a complexity of $\mathcal{O}(n.k)$ to select $k$ examples from a ground-set of $n$ examples. But, solving Eq. \eqref{eq:mini} to select $P$ mini-batches of size $k/P$ from random subsets of size $r$ has a total complexity of $\mathcal{O}(P\times r\cdot\frac{k}{P})=\mathcal{O}(r\cdot k)$.

\textbf{Remark.} Early in training, quadratic approximations are accurate in a small neighborhood $\N_l$. Hence, a smaller number of mini-batches can be extracted simultaneously. In the final stage of training, the loss can be well captured by a quadratic function \cite{martens2015optimizing}. Hence, a larger number of mini-batches can be selected simultaneously later in training. In practice, simply increasing $P$ proportional to the inverse of the norm of the Hessian diagonal works well, as we confirm by our experiments.

\subsection{Further Improving Efficiency of Coreset Selection}
To address the challenge \textbf{(C3)} of further improving the efficiency and performance of selecting coresets, we make the following observation. Examples are gradually learned during the training. When an example is learned, its gradient and loss become nearly zero. %
Hence, such examples do not affect training and can be 
dropped from the coreset selection pipeline to improve efficiently. %
However, the gradient or loss of an example at a single point during training can be very noisy. To quickly identify such examples, we monitor the loss of examples within non-overlapping intervals of length $T_2$ during the training, and exclude those that consistently have a loss smaller than a threshold $\alpha$. %
This shrinks the size of the %
selection problem
over time, and allows \alg to focus more on examples that are not learned. Hence, it further improves the efficiency and speedup of the algorithm.

To efficiently exclude the learned examples, we only rely on the loss values calculated for random subsets used for selecting the coresets, and drop examples for which the calculated loss values are smaller than $\alpha$ in an interval of length $T_2$.\looseness=-1

Effectively, dropping the learned examples speeds up training by increasing the learning rate. Specifically, dropping $s$ examples with nearly-zero gradients from a ground-set of $n$ examples increases the full (average) gradient %
by nearly ${n}/{(n-s)}$, which has a similar effect to that of increasing the learning rate by a factor of nearly ${n}/{(n-s)}$.

The pseudocode of \alg is illustrated in Alg. \ref{alg}.

The following Theorem shows that training with stochastic gradient descent on mini-batch coresets found by \alg converges to a stationary point of the non-convex loss.

\begin{theorem}\label{thm}
For any $\delta, \lambda > 0$, assume that the function $\LL$ is $L$-gradient Lipschitz, %
and stochastic gradients $\g_{t,i}$ have a bounded variance, i.e., $\E_{i\in V}[\|\g_{t,i}-\nabla \LL(\w_t)\|^2]\leq\sigma^2$. 

\textbf{Case 1 (\alg: Nearly-unbiased).} 
Let step size %
be $\eta=\min\{\frac{1}{L},\frac{\tilde{D}\sqrt{r}}{\sigma \sqrt{N}}\}$, for some $\tilde{D}\!\!>\!\!\!0$ and $N$ be the number of training iterations.
If the %
gradient bias of mini-batch coresets 
$\E[\|\xib_{t_l}\|]\!\leq\! \epsilon \|\nabla \LL(\w_{t_l})\|$
and $\tau\leq \min_l (\|\nabla\LL(\w_{t_l}+\del_l)\|-3\epsilon \|\nabla \LL(\w_{t_l})\|)\|\del_l\|/2\LL(\w_{t_l}\!+\del_l)$, %
for $0\leq\!\epsilon\!\leq\! \min\{\!1,\!\|\nabla\LL(\w_{t_l}\!+\del_l)\|/3\|\nabla \LL(\w_{t_l})\|\}$, then with probability at least $1 -\lambda$, \alg will visit a $\nu$-stationary point at least once in the following number of iterations: \looseness=-1
\begin{equation}
    \tilde{\mathcal{O}}\left(\frac{L(\LL(\w_0)-\LL^*)}{{\nu}^2}(1+\frac{\sigma^2}{{r\nu}^2})\right).
\end{equation}
\textbf{Case 2 (Biased).}
If %
the bias of mini-batches
$\E[\|\xib_t\|]\leq \epsilon$, but $\epsilon$ is larger than the full gradient norm anytime during the training, %
then the number of iterations is:\looseness=-1
\begin{align}\label{eq:biased}
    \tilde{\mathcal{O}}\left(\frac{L(\LL(\w_0)-\LL^*)}{\nu^2-\epsilon}(1+\frac{\sigma^2+r\epsilon^2}{r(\nu^2-\epsilon)})\right).
\end{align}
In particular, if $\epsilon\geq \nu^2$, convergence is not guaranteed.
\end{theorem}
\begin{table*}[th!]
\caption{Relative error (\%) of different methods over training on the full data. All the baselines select subsets of size 10\% of full data at the beginning of every epoch. On the other hand, \alg selects mini-batches and 
decides when to update the mini-batches based on its quadratic loss approximation. (*) \glister uses the validation set, and (\ddag) \grad uses higher dimensional gradients to find coresets. SGD\textdagger\ shows accuracy of a standard mini-batch SGD pipeline at 10\% training.\looseness=-1
}
\label{tab:baseline}
\vspace{-4mm}
\begin{center}
\begin{small}
\begin{sc}
\resizebox{\textwidth}{!}{%
\begin{tabular}{lcrrrrrr}
\toprule
Dataset ~-~ Model & Backprop & SGD\textdagger & Random & Craig & Grad-match\ddag & Glister* &  \textbf{\alg} (Ours)\\
\midrule
CIFAR-10 ~-~ ResNet-20 & 10\% & $21.3_{\pm 8.0}$ & $7.2_{\pm 1.4}$ & $13.0_{\pm 5.1}$ & $6.0_{\pm 0.1}$ & $7.0_{\pm 0.1}$ & $\mathbf{5.5_{\pm 0.2}}$  \\
CIFAR-100 ~-~ ResNet-18 & 10\% & $36.5_{\pm 2.9}$ & $11.7_{\pm 0.4}$ & $17.2_{\pm 4.5}$ & $12.7_{\pm 0.9}$ & $27.6_{\pm 4.0}$ & $\mathbf{9.4_{\pm 0.3}}$  \\
TinyImageNet ~-~ ResNet-50 & 10\% & $32.8_{\pm 2.1}$ & $16.0_{\pm 0.5}$ & $28.5_{\pm 0.6}$ & $27.7_{\pm 0.2}$ & $32.8_{\pm 2.1}$ & $\mathbf{15.4_{\pm 0.6}}$ \\
SNLI ~-~ RoBERTa (Finetune) & 10\% & $1.2_{\pm 0.3}$ & $1.2_{\pm 0.3}$ & - & - & - & $\mathbf{0.8_{\pm 0.2}}$ \\
\bottomrule
\end{tabular}%
}
\end{sc}
\end{small}
\end{center}
\vskip -0.2in
\end{table*}
The proof can be found in Appendix \ref{appendix:proofs}.
At a high level, Theorem \ref{thm} shows that if mini-batch coresets closely capture gradient of random subsets $V_p$, \alg with a small enough $\tau$, converges to a $\nu$-stationary point of the non-convex loss, but $r/m$ times faster than mini-batch SGD with mini-batch size $m$ on full data, as discussed next. %

\textbf{Case 1.} 
As \alg mini-batch coresets capture the gradient of random subsets closely, the bias of mini-batch coresets is a small fraction, $\epsilon\in[0,1]$, of the full gradient norm at selection time. %
If %
$\epsilon\leq \min\{1,\|\nabla \LL(\w_{t_l}\!+\!\del_l)\|/3\|\nabla \LL(\w_{t_l})\|\}$,
a small enough $\tau$ ensures that %
the bias
stays smaller than the full gradient norm %
within the neighborhood $\N_l$ (\textit{c.f.} \cref{fig:bias_normed} in Appendix \ref{appendix:details}). %
Importantly, as the gradient norm shrinks as we get close to a stationary point, a small $\epsilon$ implies that the bias in the entire neighborhood vanishes close to convergence.
This guarantees convergence of \alg to a $\nu$-stationary point. Notably, as long as $r\leq \sigma^2/\nu^2$, training with \alg linearly speeds up training by a factor of $r$. In particular, compared to SGD with mini-batch size $m$, \alg speeds up training by a factor of $r/m$. \looseness=-1

\textbf{Case 2.} \hspace{-2mm} If the bias of the mini-batch gradients $\epsilon$ is %
is larger than the full gradient norm or larger 
than $\nu$, (mini-batch) SGD does not converge to a $\nu$-stationary point. %
This explains why larger bias of mini-batches selected from coresets extracted from the full data results in a poor performance (\textit{c.f.} \cref{fig:fail_acc}). Besides, the $\epsilon$ bias slows down the training by a factor of $\nu^2\!-\epsilon$. Note that such mini-batches also have a larger variance than mini-batch coresets found by \alg, which should be replaced by $1/r$ in Eq. \eqref{eq:biased}.

\vspace{-1mm}
\section{Experiments}
\vspace{-1mm}
In this section, we evaluate the performance of our coreset selection, \alg.
{First}, we compare \alg to the state-of-the-art coreset selection algorithms, namely \craig \cite{mirzasoleiman2020coresets}, \glister \cite{killamsetty2021glister}, and \grad \cite{killamsetty2021grad}, as well as the Random baseline. 
Second, we evaluate the effectiveness of our quadratic approximations in determining the time that %
the coresets needs to be updated.
In addition, we compare the speedup of training with \alg to other baselines.
Then, we conduct an ablation study to investigate the necessity of the quadratic vs. linear approximation, smoothing gradient and curvature, and dropping the learned examples from the selection pipeline. 
Finally, we study the learning difficulty of subsets that are selected by \alg during the course of training.

\textbf{Datasets and Models.}
To demonstrate the effectiveness of \alg across different datasets and architectures, we apply \alg to several image and language benchmarks, including
training ResNet-20 on CIFAR10, ResNet-18 on CIFAR-100 \cite{cifar10}, ResNet-50 on TinyImageNet \cite{imagenet15russakovsky}, and fine-tuning RoBERTa on Stanford Natural Language Inference (SNLI) \cite{bowman-etal-2015-large}. 
\cref{tab:dataset} summarizes the datasets and models.\looseness=-1

\textbf{Training Setup.} For all datasets except SNLI, we consider a standard deep learning training pipeline that runs for 200 epochs with a %
SGD optimizer with a momentum of 0.9, and %
decays the learning rate by a factor of 0.1 after 60\% and 85\% of training, and use mini-batch size 128. 
We warm-start the learning rate to 0.1 in the first 10\% of training, which is essential for stability of all the methods, except \alg and Random. However, for fair comparison, we compare all the methods using learning rate warm-start. For fine-tuning RoBERTa on SNLI  we used an AdamW optimizer and a learning rate of 1e-5 for 8 epochs, with mini-batch size 32.
We ran all experiments with a single NVIDIA RTX A6000 GPU.  \looseness=-1 %

\textbf{Evaluation.} We evaluate all the methods under 10\% budget for training. That is, for \craig, \grad, and \glister, we find a new coreset of size 10\% of the full data at the beginning of \textit{every epoch}. On the other hand, Random iteratively selects random mini-batches, and \alg finds mini-batch coresets and automatically finds the time to update them. We stop all methods after the same number of training iterations as that of 10\% training on the full data.
Note that under the above `training setup', the Random baseline achieves a much higher accuracy than that of epoch 20 of a standard 200 epoch training pipeline (see SGD\textdagger\ in Table \ref{tab:baseline}). This is because %
the learning rate drops twice during training on Random (and coresets) under 10\% budget.

\begin{figure*}[t]
\begin{center}
  \begin{subfigure}[t]{0.24\textwidth}
     \centering
     \includegraphics[trim={7mm 0 7mm 0},clip, width=\columnwidth]{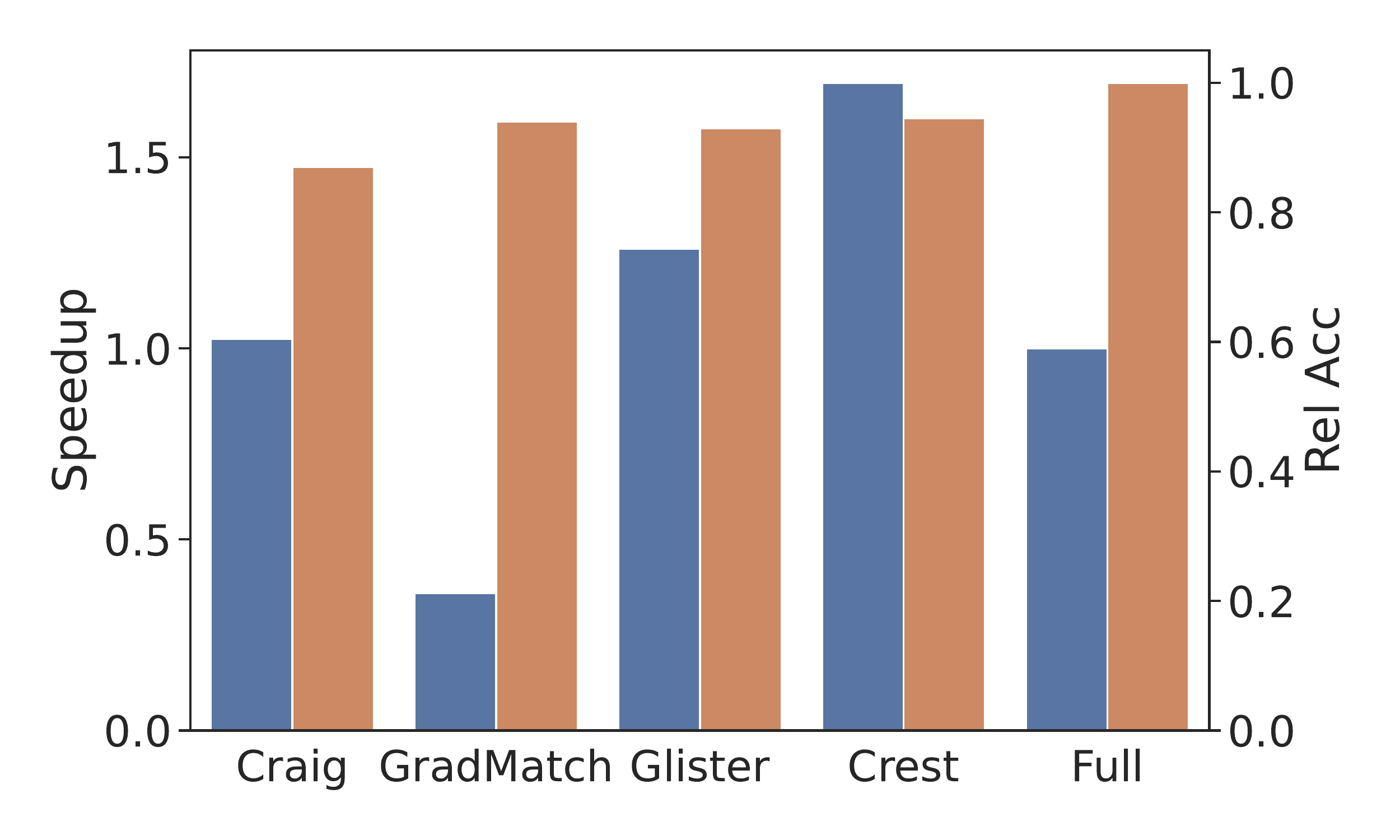}
     \caption{CIFAR-10}
     \label{fig:cifar10-speedup}
 \end{subfigure}
 \begin{subfigure}[t]{0.24\textwidth}
     \centering
     \includegraphics[trim={7mm 0 7mm 0},clip, width=\columnwidth]{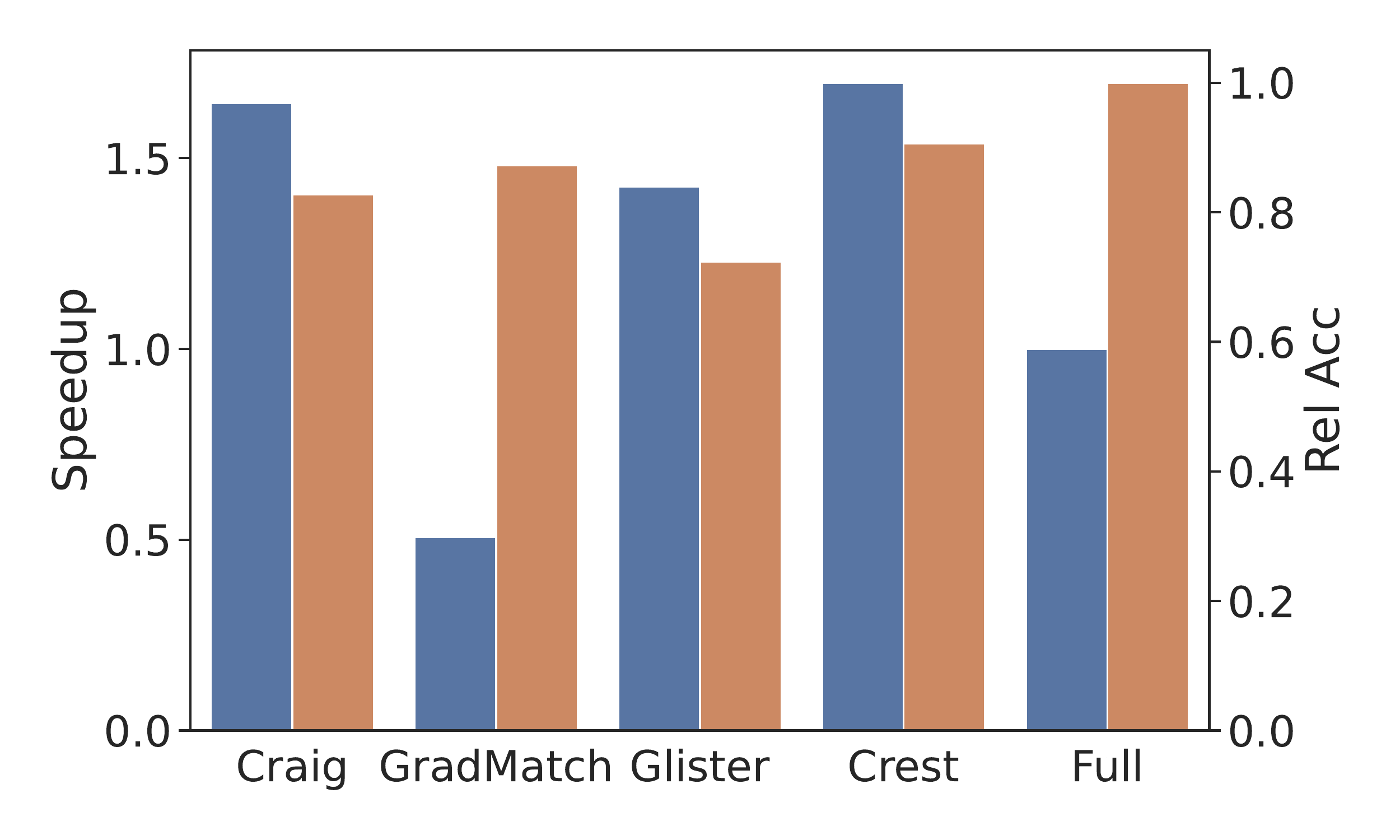}
     \caption{CIFAR-100}
     \label{fig:cifar100-speedup}
 \end{subfigure}
 \begin{subfigure}[t]{0.24\textwidth}
     \centering
    \includegraphics[trim={7mm 0 7mm 0},clip, width=\columnwidth]{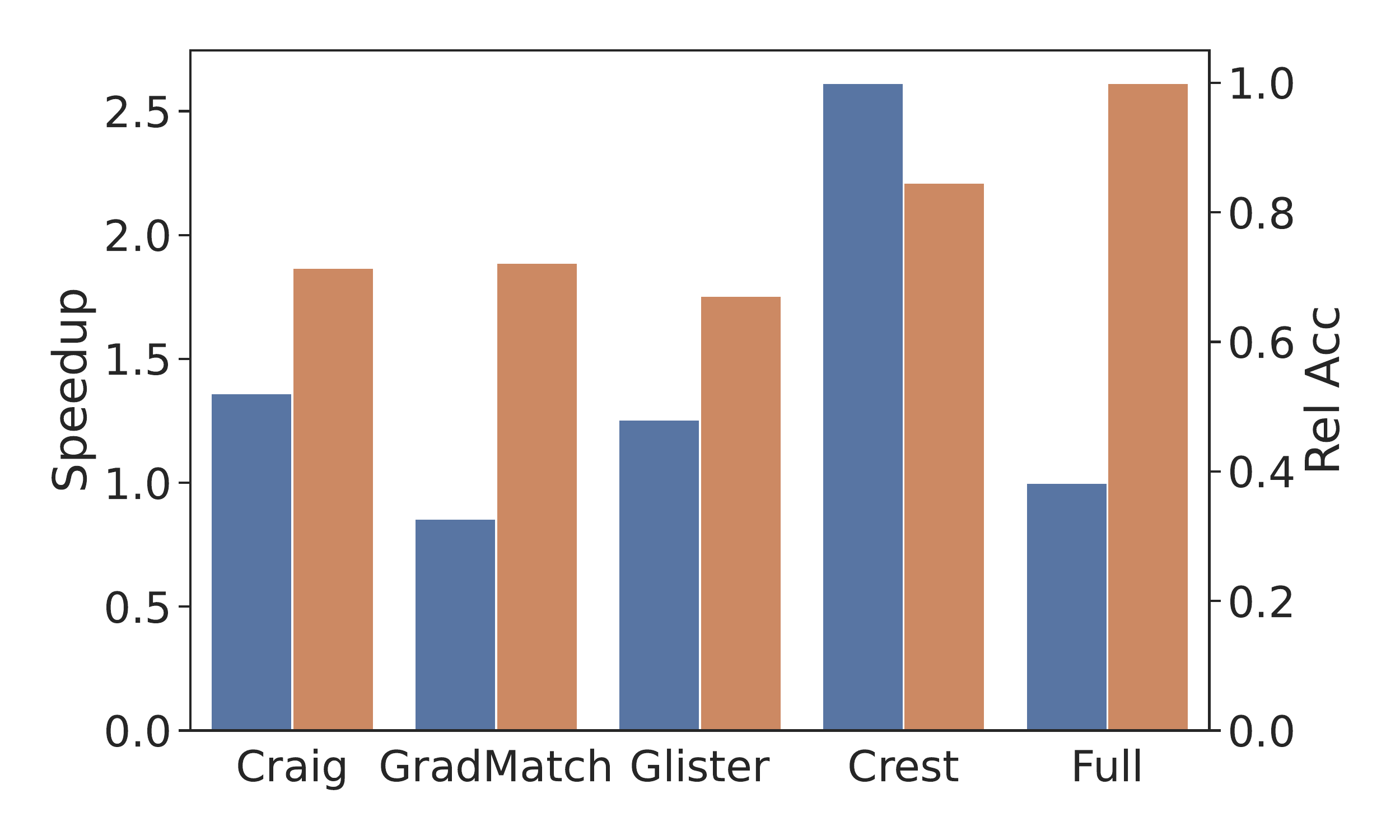}
     \caption{TinyImageNet}
     \label{fig:tinyimagenet-speedup}
 \end{subfigure}
 \begin{subfigure}[t]{0.24\textwidth}
     \centering
    \includegraphics[trim={7mm 0 7mm 0},clip, width=\columnwidth]{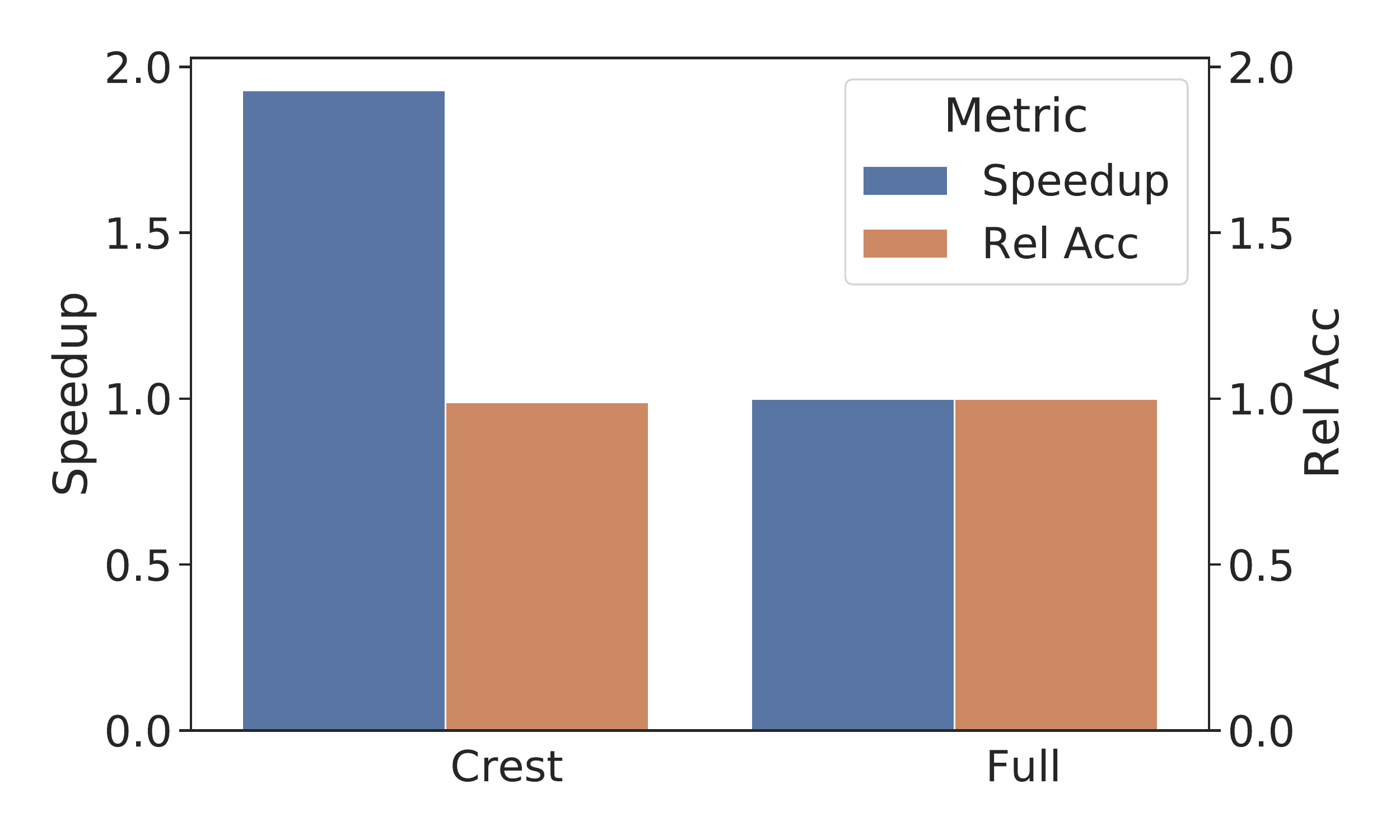}
    \vspace{-5mm}
     \caption{SNLI}
     \label{fig:snli-speedup}
 \end{subfigure}
\vspace{-3mm}
\caption{Normalized run-time and test accuracy of \alg by that of full data, when training ResNet-20 on CIFAR10, ResNet-18 on CIFAR100, ResNet-50 on TinyImagenet, and fine-tuning RoBERTa on SNLI. 
}
\label{fig:runtime}
\end{center}
\vspace{-6mm}
\end{figure*}

\textbf{\alg Setup.} %
In our experiments, we used $b=5$, and $T_2=20$ for all the datasets, and tuned $\tau, \alpha$ and $h$, as discussed in Appendix \ref{appendix:details}.
Nevertheless, our method is not very sensitive to the choice of $\alpha$. We used $|V_p|=|V_r|=r=0.005\times n$ for SNLI and $0.01\times n$ for the rest of the datasets, without further tuning.
For RoBERTa we used last layer gradient and Hessian diagonal, and for other networks we used full gradient and Hessian diagonal in Eq. \eqref{eq:m}. \looseness=-1

\subsection{Evaluating Accuracy and Speedup}
\label{sec:sgd-coreset}
\textbf{Accuracy.} \cref{tab:baseline} shows the relative error, i.e., $\frac{|acc_{coreset} - acc_{full}|}{acc_{full}}$ of models trained with each coreset selection algorithm. 
We see that while the baselines yield a very high relative error in particular for larger models and more difficult tasks, e.g. CIFAR100 and TinyImageNet, \alg can successfully outperform all the baselines and obtain up to 18.2\% better relative error compared to baseline coreset selection methods, and up to 2.3\% better relative error compared to Random baseline. 
Note that as the size of the data increases, existing methods that select coresets from the full data become prohibitively expensive. Notably, \alg is the only coreset selection method that is applicable to SNLI with 570k examples. Other coreset baselines that find subsets from the full data cannot scale to such a large data. 
\cref{tab:baseline} confirms that \alg can successfully finds mini-batch coresets with small bias and variance and identify when they need to be updated during the training.

\textbf{Speedup.} 
\cref{fig:runtime} compares the accuracy and wall-clock run time of \alg vs  baselines, and training on full data. We see that \alg is able to achieve up to 2.5x speeds up over training on full data, while introducing the smallest relative error compared to the baselines, when training ResNet-20 on CIFAR-10, ResNet-18 on CIFAR-100, ResNet-50 on TinyImageNet, and fine-tuning RoBERTa on SNLI. 
\cref{tab:run} further lists the average wall-clock time for selecting every mini-batch coreset of size 128, %
calculating the quadratic loss approximation based on Eq. \eqref{eq:m}, and checking the validity of the approximation on a random subset of data according to Eq. \eqref{eq:threshold}, when selecting coresets with \alg to train ResNet-18 on CIFAR100.
Note that selecting a mini-batch from a larger random subset is much faster than selecting a subset of size 10\% from the full data, done by the baselines. %

\subsection{Ablation Study}
\begin{figure}[t]
\begin{center}
\begin{subfigure}[t]{0.24\columnwidth}
     \centering
     \includegraphics[height=1.24\columnwidth]{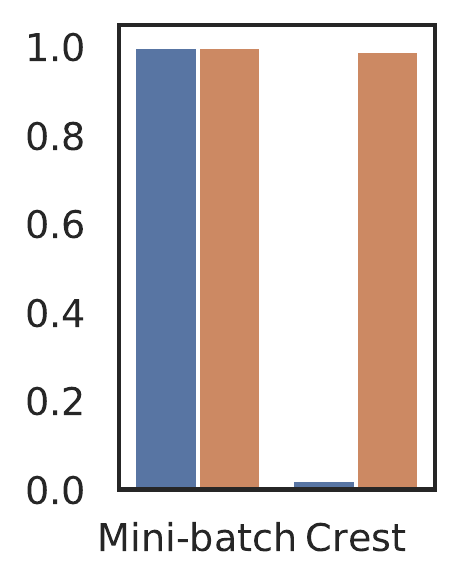}
     \caption{\scriptsize{CIFAR-10}}
     \label{fig:cifar10-minibatch}
 \end{subfigure}
 \begin{subfigure}[t]{0.24\columnwidth}
     \centering
     \includegraphics[height=1.24\columnwidth]{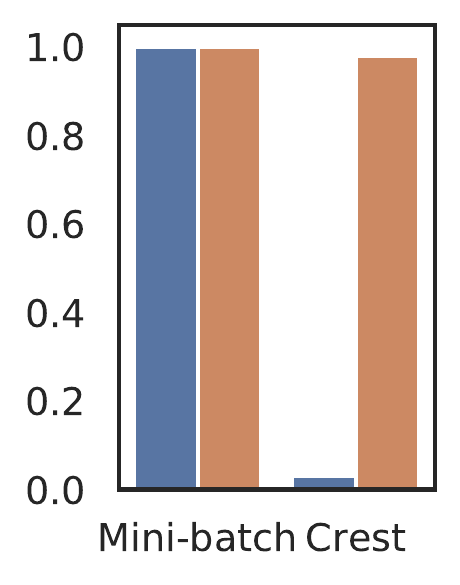}
     \caption{\scriptsize{CIFAR-100}}
     \label{fig:cifar100-minibatch}
 \end{subfigure}
 \begin{subfigure}[t]{0.24\columnwidth}
     \centering
     \includegraphics[height=1.24\columnwidth]{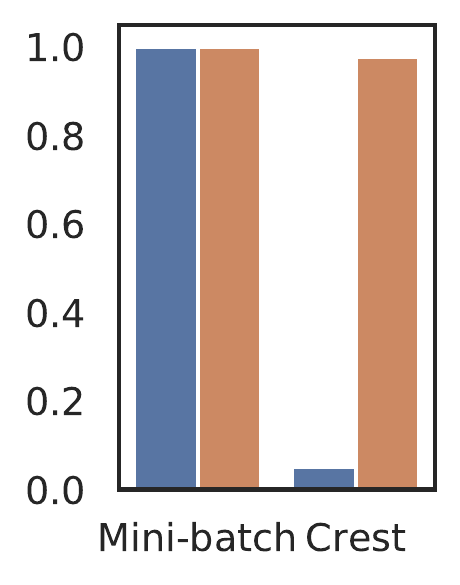}
     \caption{\scriptsize{TinyImageNet}}
     \label{fig:tinyimagenet-minibatch}
 \end{subfigure}
 \begin{subfigure}[t]{0.24\columnwidth}
     \centering
    \includegraphics[height=1.24\columnwidth]{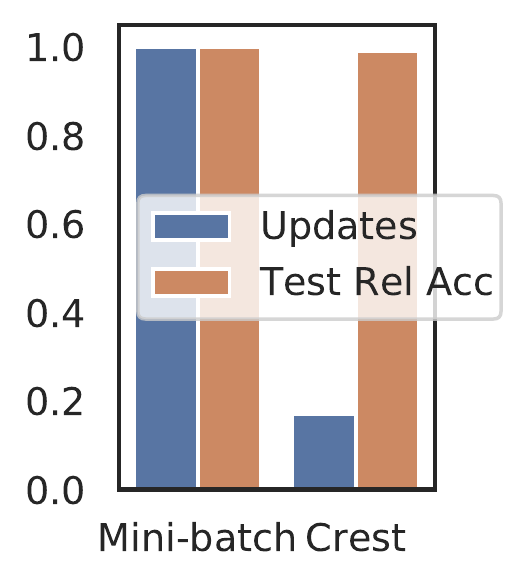}
     \caption{\scriptsize{SNLI}}
     \label{fig:snli-minibatch}
 \end{subfigure}
\vspace{-2mm}
\caption{Normalized test accuracy and number of coreset updates for \alg over greedily selecting every mini-batch from a larger random subset by solving Eq. \eqref{eq:sub_max}. %
}
\label{fig:acc}
\end{center}
\vspace{-7mm}
\end{figure}

\textbf{Modeling the loss.}
\label{sec:approx-coreset}
Next, we evaluate the effectiveness of \alg in approximating the loss as piece-wise quadratic regions and identifying the time that the coresets need to be updated. 
To do so, we compare \alg with greedy mini-batch selection, which selects every mini-batch by applying the greedy algorithm to solve Eq. \eqref{eq:sub_max} on one random subset, and trains on it before selecting the next mini-batch.
\cref{fig:acc} compares the relative error and the number of times \alg updates the coresets to greedy mini-batch selection. 
We see that \alg can effectively reduce the number of updates to 2\% and 3\% of the total update time of greedy mini-batch selection while preserving 98\% and 99\% of its performance, when training ResNet18 on CIFAR-100 and ResNet20 on CIFAR-10. For training ResNet50 on TinyImagenet and fine-tuning RoBERTa on SNLI, \alg reduces the number of updates to 19\% and 26\% respectively, while preserving 95\% and 99\% of the performance.\looseness=-1

\label{sec:ablation}
\begin{table}[t]
\caption{Average time for different components of \alg for training ResNet-18 on CIFAR-100 with batch size 128. 
}
\label{tab:run}
\vspace{-3mm}
\begin{center}
\begin{small}
\begin{sc}
\begin{tabular}{lc}
\toprule
Step & Time (seconds) \\
\midrule
selection (\alg) & 0.006\\
selection (\craig) & 0.089\\
Loss approximation & 0.115 \\
Checking threshold & 0.796\\
\bottomrule
\end{tabular}
\end{sc}
\end{small}
\end{center}
\vspace{-8mm}
\end{table}

\begin{table}[t]
\caption{\hspace{-1mm}Effect of \alg components  (ResNet20/CIFAR10).\looseness=-1
}
\label{tab:ablation}
\vspace{-3mm}
\begin{center}
\begin{small}
\begin{sc}
\begin{tabular}{lcc}
\toprule
Algorithm & Rel. Error & \# Updates\\
\midrule
\alg-First & 7.45 & 343 \\
\alg w/o smooth & 7.44 & 369 \\
\alg w/o excluding & 4.61 & 346\\
 \alg & 4.33 & 185 \\
\bottomrule
\end{tabular}
\vspace{-4mm}
\end{sc}
\end{small}
\end{center}
\end{table}

\begin{figure}[ht]
\begin{center}
\centerline{\includegraphics[trim={7mm 0 5mm 0},clip, width=0.5\columnwidth]{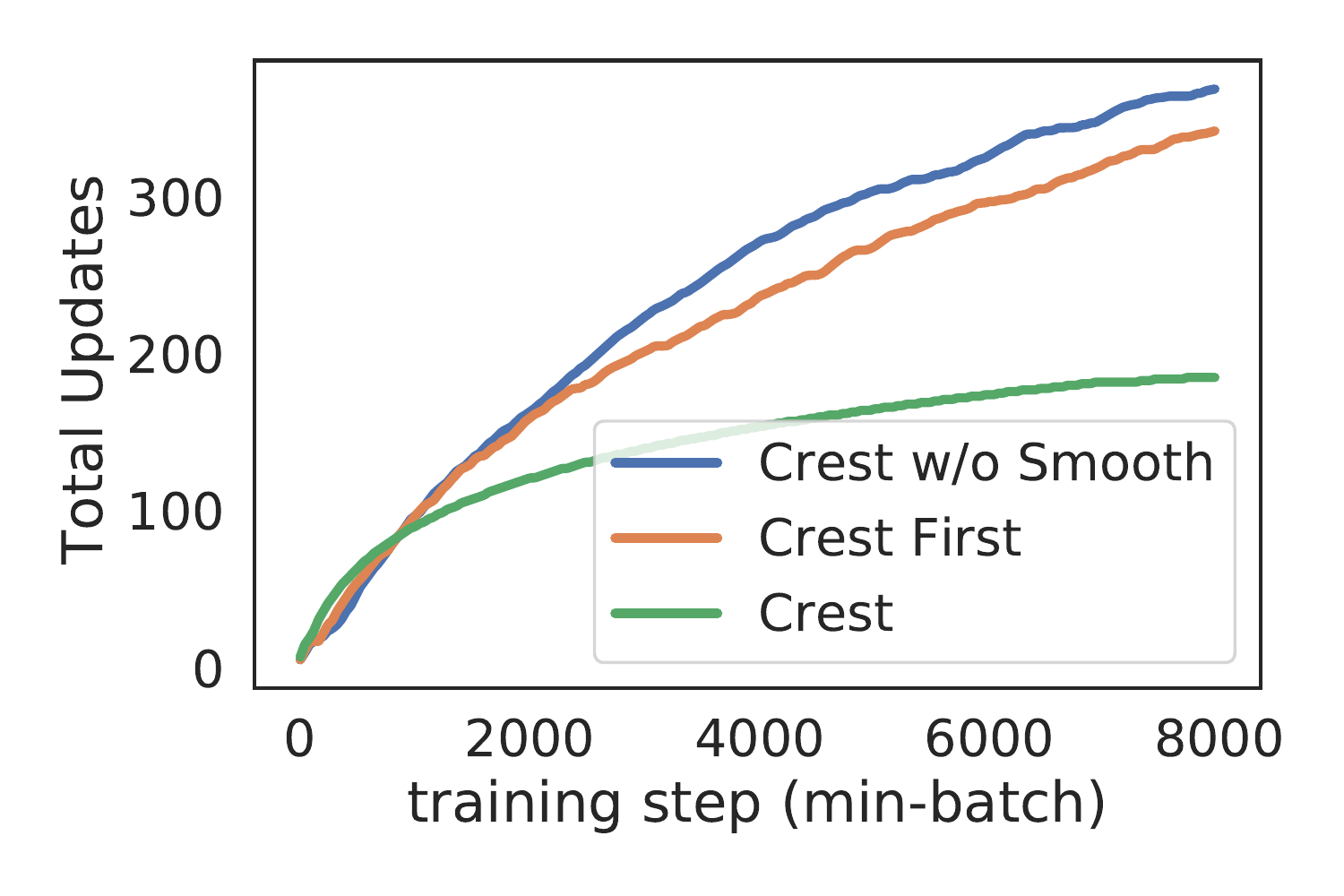}\includegraphics[trim={7mm 0 5mm 0},clip, width=0.5\columnwidth]{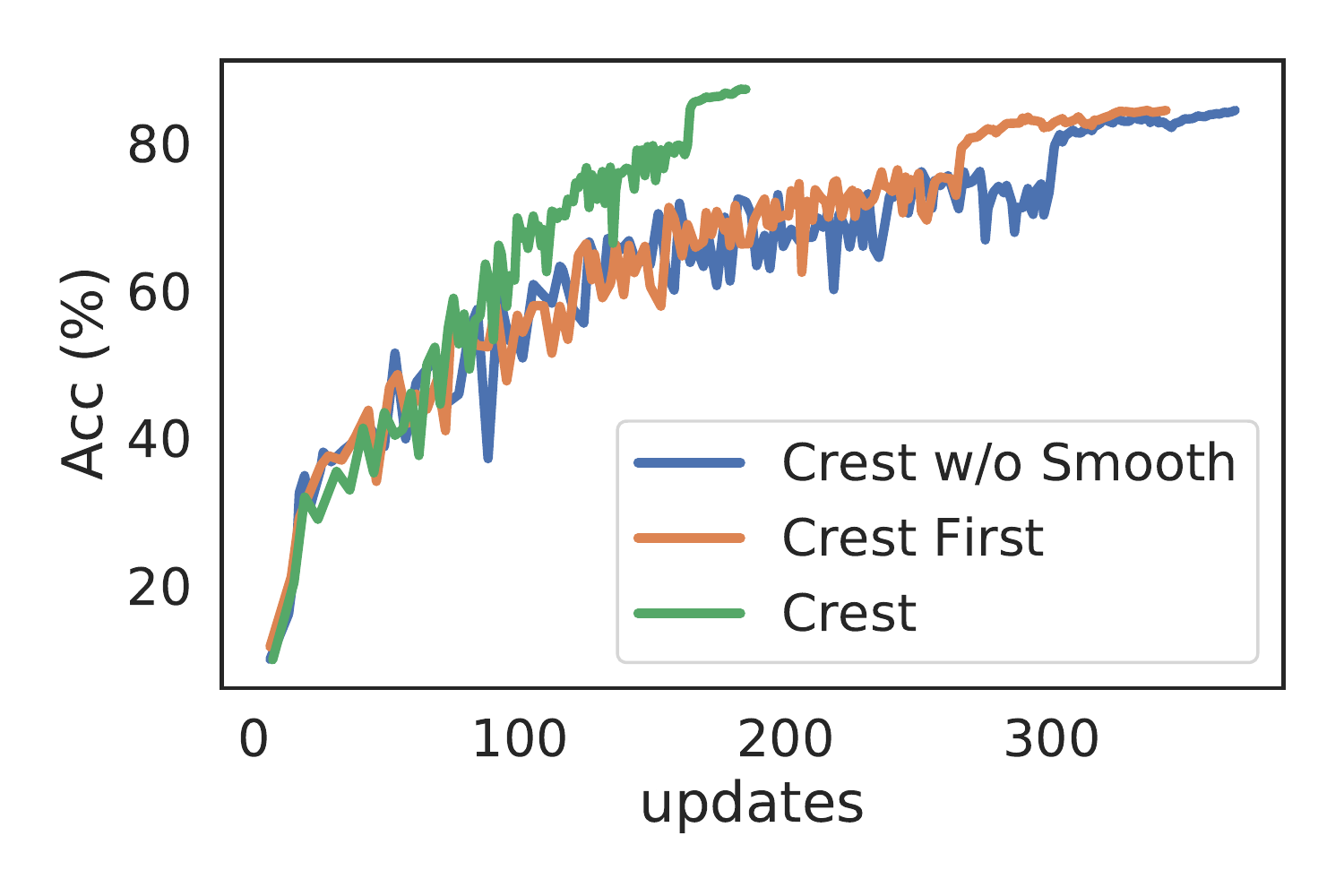}}
\vspace{-3mm}
\caption{Training ResNet-20 on CIFAR-10 with \alg %
under 10\% training budget. (Left) Number of coreset updates vs. training iterations. (Right) test accuracy vs. the total number of coreset updates. \vspace{-2mm}}
\label{fig:ablation}
\end{center}
\vskip -0.3in
\end{figure}

\textbf{Quadratic approximation.}
As discussed in Sec. \ref{sec:model}, in later stages of training, the loss can be better approximated as a convex quadratic function within larger neighborhoods.
\cref{fig:ablation} (left) shows that as training proceeds, \alg can successfully increase the size of the neighborhoods in which the quadratic approximation is valid, and reduce the number of updates over time.
Moreover, \cref{fig:ablation} (right) shows that using a first-order approximation instead of our quadratic approximation, or not smoothing the gradient and curvature in calculating the quadratic approximation, leads to higher number of coreset updates, and harms the accuracy. %
\cref{tab:ablation} further compares the number of updates and the relative error at the end of training.
We see that excluding the learned examples further improves the performance of \alg. %

\cref{tab:ablation} and \cref{fig:ablation} show that it is crucial \textit{when} the coresets are updated. \cref{fig:ablation} shows that updating the coreset more frequently in the beginning is the key (notice that the green line is slightly higher than blue and orange in the first 1000 iterations). This slight difference results in a much better final accuracy. However, updating the coreset frequently later in training does not improve the accuracy (it does not hurt but does not help). Hence, blue and orange lines achieve a lower accuracy than Crest with more updates. \alg can accurately find when is best to update the coresets based on its quadratic loss approximation, and achieve a better accuracy while minimizing the number of updates.

\begin{figure}[h]
\begin{center}
\centerline{\includegraphics[trim={5mm 0 5mm 0},clip,width=0.5\columnwidth]{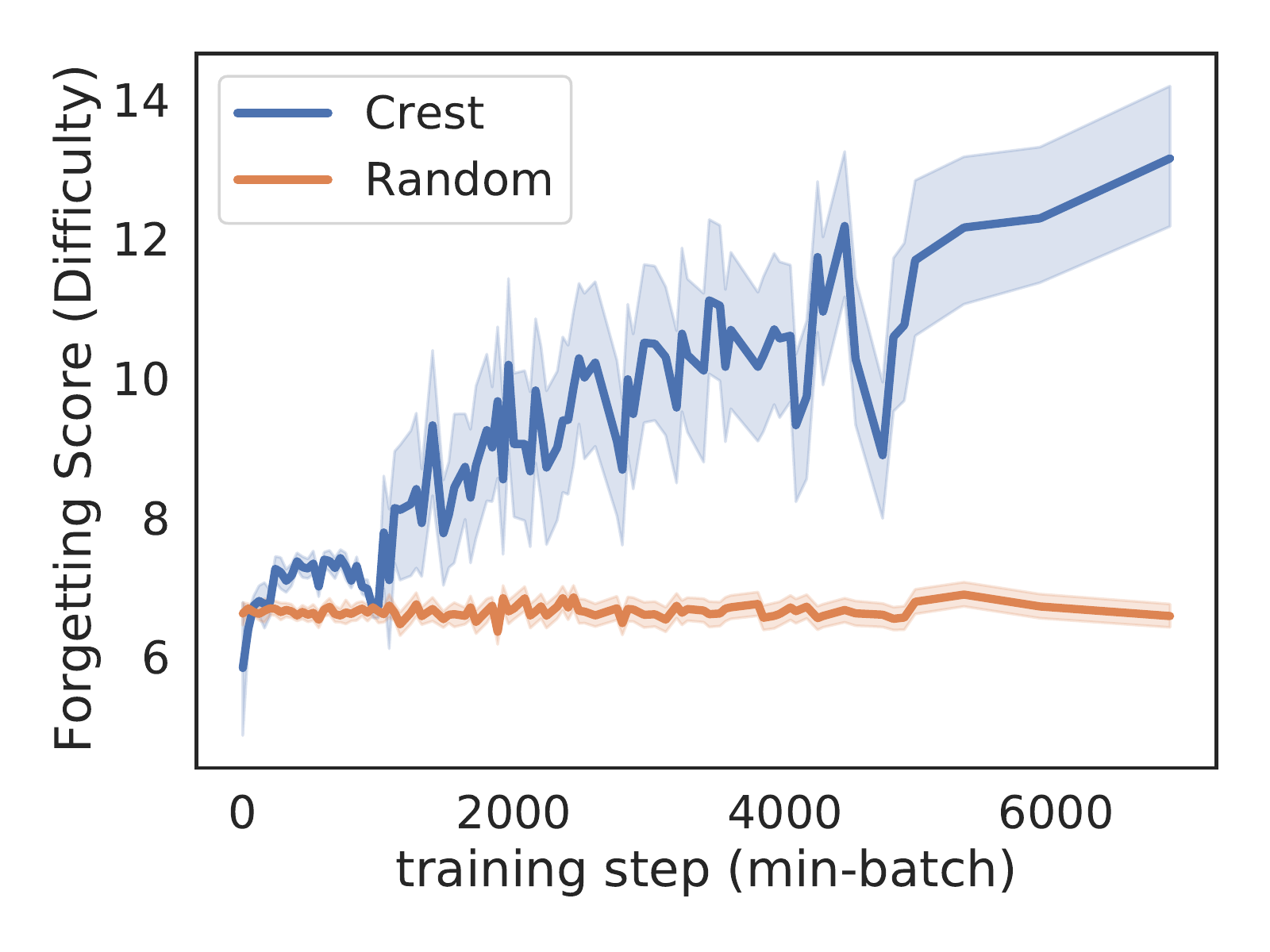}\includegraphics[trim={5mm 0 5mm 0},clip,width=0.5\columnwidth]{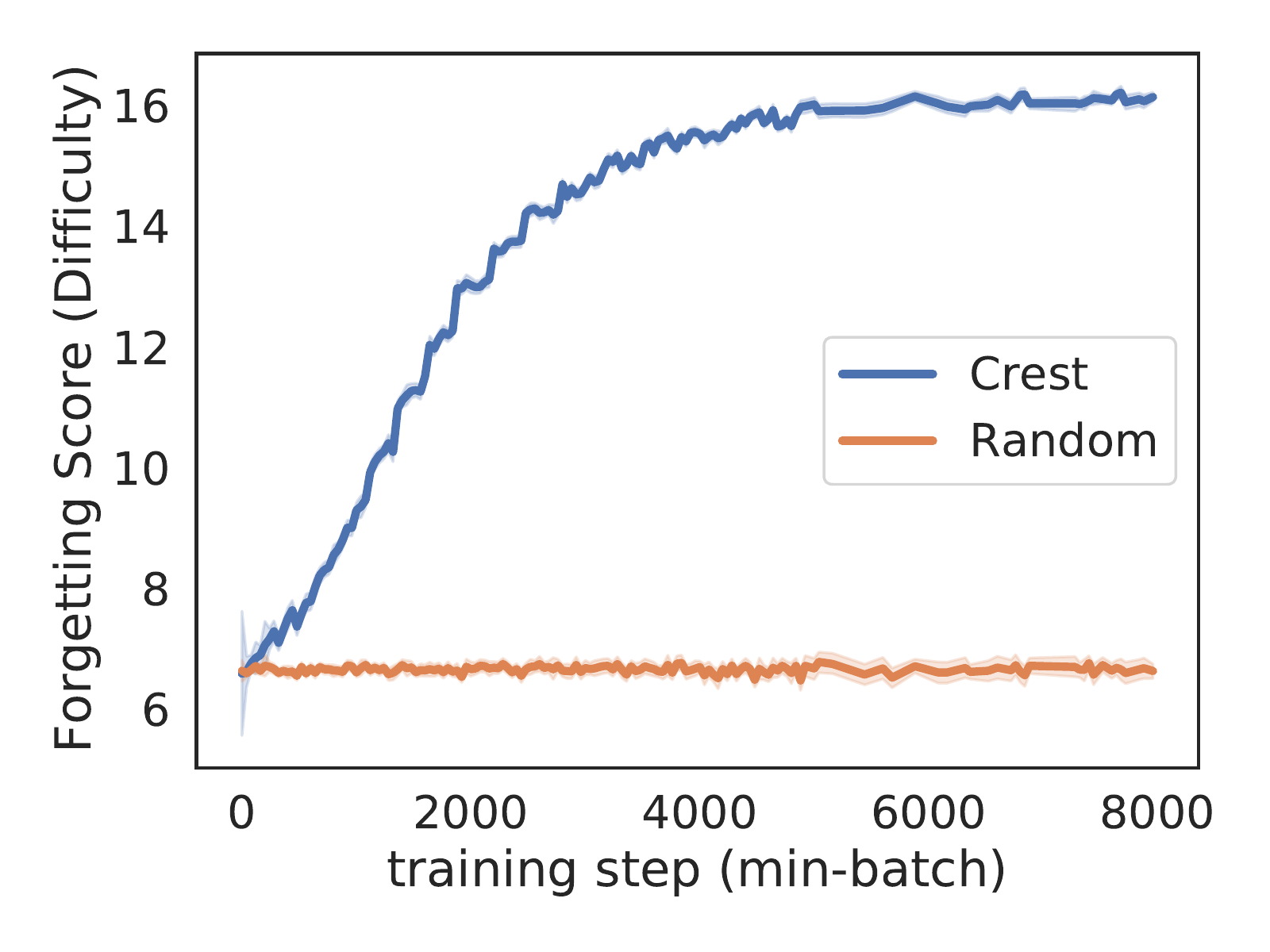}}
\vspace{-3mm}
\caption{Average forgettability score of \alg coresets during training, when learned examples are not discarded (Left), and are discarded (Right). Learning difficulty of examples selected by \alg increases during the training.\looseness=-1 %
\vspace{-2mm}}
\label{fig:curriculum}
\end{center}
\vspace{-9mm}
\end{figure}

\textbf{Importance of Examples during the Training.} 
\cref{fig:curriculum} shows the average forgetting score for the selected examples during the training. Forgetting score counts the number of times examples are misclassified after being correctly classified during the training, and quantifies the difficulty of learning an example \cite{toneva2018empirical}. We see that \alg selects examples of increasing difficulty during the training, and excluding the learned examples allows further focusing on the difficult-to-learn examples. %
In contrast, random subsets have a constantly lower forgetting score during the training. 
\cref{fig:forget} in Appendix \ref{appendix:details} shows that while \alg trains on a diverse set of examples, the distribution of the number of times different examples are selected by \alg is very long-tailed. %
This shows not all examples contribute equally to training, and \alg can successfully find examples that are important for learning at different times. \looseness=-1 %

\textbf{Limitations.} In general,
coreset methods are most beneficial under a limited training budget. While \alg can still achieve a superior accuracy %
under a larger budget (\cref{tab:20-acc} in Appendix \ref{appendix:details}), it achieves a smaller accuracy gap %
compared to the Random baseline. Besides, more efficient data loading can significantly speed up coreset selection.

\vspace{-1mm}
\section{Conclusion}
\vspace{-1mm}
We proposed the first scalable framework with rigorous theoretical
guarantees to identify the most valuable examples for training non-convex models, particularly
deep networks. Our approach models the non-convex loss as a series of quadratic functions and extracts a coreset for each quadratic sub-region. In addition, to ensure convergence of stochastic gradient methods such as (mini-batch) SGD, it iteratively extracts multiple coresets from smaller random subsets of training data, to ensure nearly-unbiased gradient estimates with small variance.
In doing so, it provides rigorous theoretical guarantee for convergence of the extracted coresets to stationary point of a non-convex function. With extensive experiments, we confirmed the effectiveness of our method on various vision and NLP deep learning tasks.

\textbf{Acknowledgment.}
This research was supported by %
the National Science Foundation CAREER
Award 2146492.\looseness=-1

\bibliography{piece}
\bibliographystyle{icml2023}

\onecolumn
\appendix
\section{Appendix}
\subsection{Proofs}\label{appendix:proofs}
We assume that the stochastic gradients are unbiased and have a bounded variance, i.e., 
\begin{equation}
    \E_{i\in V}[\|\g_{t,i}-\g_{t,V}\|]=\E[\|\zetab_t\|]=0, \quad \quad \E_{i\in V}[\|\g_{t,i}-\g_{t,V}\|^2]=\E[\|\zetab_t\|^2]\leq \sigma^2.
\end{equation}
Also assume that the function
$\LL$ is $L$-gradient Lipschitz, i.e., 
\begin{align}
    \|\nabla \LL(\w_1)-\nabla \LL(\w_2)\|\leq L\|\w_1-\w_2\|, \quad \forall \w_1,\w_2\in \mathcal{W}.
\end{align}
Then, we have that:
\begin{align}\label{eq:lipschitz}
    |\LL(\w_1)-\LL(\w_2)-\left<\nabla \LL(\w_2),\w_1-\w_2\right>|\leq \frac{L}{2}\|\w_1-\w_2\|^2, \quad \forall \w_1,\w_2\in \mathcal{W}.
\end{align}
We can write the gradient descent updates when training on mini-batch coresets found by \alg, as follows:
\begin{align}\label{eq:sgd}
&\w_{t+1} \leftarrow \w_t- \eta_t (\nabla \LL(\w_t)+\tilde{\zetab}_t), \quad\quad s.t. \quad\quad \tilde{\zetab}_t=\zetab_t+\xib_t
\end{align}
where $\zetab_t$ is the error of random subset $V_p$ in capturing the full gradient, and $\xib_t$ is the error of mini-batch coreset $S_l^p$ in capturing the gradient of $V_p$. 

We build on the analysis of \cite{ghadimi2013stochastic} 
and characterize the effect of the coreset gradient error on the convergence.
From Eq. \eqref{eq:lipschitz}, \eqref{eq:sgd} we have: %
\begin{align}
\LL(\w_{t+1}) 
&\leq \LL(\w_t) + \left<\nabla \LL( \w_t ) , \w_{t+1} - \w_t \right> + \frac{L}{2} \eta_t^2 \|\nabla \LL(\w_t)+\tilde{\zetab}_t\|^2\\
& \leq \LL(\w_t) -\eta_t \left<\nabla \LL( \w_t ) , \nabla \LL(\w_t)+\tilde{\zetab}_t \right> + \frac{L}{2} \eta_t^2 \|\nabla \LL(\w_t)+\tilde{\zetab}_t\|^2\\
& = \LL(\w_t) -\eta_t \|\nabla \LL(\w_t)\|^2 - \eta_t \left< \nabla \LL ( \w_t ) , \tilde{\zetab_t} \right > + \frac{L}{2} \eta_t^2\Big[\|\nabla \LL(\w_t)\|^2+2\left< \nabla \LL ( \w_t ) , \tilde{\zetab_t} \right >+\|\tilde{\zetab_t}\|^2\Big] \\
& = \LL(\w_t) -(\eta_t-\frac{L}{2}\eta_t^2) \|\nabla \LL(\w_t)\|^2 - (\eta_t-L\eta_t^2) \left< \nabla \LL( \w_t ) , \tilde{\zetab_t} \right > + \frac{L}{2} \eta_t^2\|\tilde{\zetab_t}\|^2 
\end{align}
For $\eta_t < 2/L$, we have $\eta_t-L\eta_t^2/2>0$. %
Summing up the above inequalities and re-arranging the terms, we obtain: 
\begin{align}
    \sum_{t=1}^{\Te}(\eta_t-\frac{L}{2}\eta_t^2) \|\nabla \LL(\w_t)\|^2 
    &\leq \LL(\w_0)-\LL(\w_{\Te+1})-\sum_{t=1}^{\Te} (\eta_t-L\eta_t^2)\left< \nabla \LL ( \w_t ) , \tilde{\zetab_t} \right > + \frac{L}{2} \sum_{t=1}^{\Te} \eta_t^2 \|\tilde{\zetab_t}\|^2 \\
    & \leq \LL(\w_0)-\LL^*-\sum_{t=1}^{\Te} (\eta_t-L\eta_t^2)\left< \nabla \LL ( \w_t ) , \tilde{\zetab_t} \right > + \frac{L}{2} \sum_{t=1}^{\Te} \eta_t^2 \|\tilde{\zetab_t}\|^2,\label{eq:l*}\\
    &= \LL(\w_0)-\LL^*
    -\sum_{t=1}^{\Te} (\eta_t-L\eta_t^2)\left< \nabla \LL ( \w_t ) , \zetab_t+\xib_t \right >
    + \frac{L}{2} \sum_{t=1}^{\Te}\eta_t^2\|\zetab_t+\xib_t\|^2,\\
    &= \LL(\w_0)-\LL^*
    -\sum_{t=1}^{\Te} (\eta_t-L\eta_t^2)\left< \nabla \LL ( \w_t ) , \zetab_t+\xib_t \right >
    + \frac{L}{2} \sum_{t=1}^{\Te}\eta_t^2( \|\zetab_t^2\|%
    +\|\xib_t^2\|+2\left<\zetab_t,\xib_t\right>),\label{eq:exp}
\end{align}
where $\LL^*$ is the optimal solution and Eq. \eqref{eq:l*} follows from the fact that $\LL (\w_{\Te+1} ) \geq \LL^*$. 
Taking expectations (with respect to the history $\Psi_{N}$ of the generated random process) on both sides of Eq. \eqref{eq:exp} and noting that $\E[\|\zetab_t\|]=0$, and $\E[\|\zetab_t\|^2]\leq \sigma^2$, and 
$\E[\left <\nabla \LL(\w_t),\zetab_t\right>|\Psi_{t-1}] = 0$, and $\mathbb{E}[\left <\zetab_t,\xib_t\right>|\Psi_{t-1}] = 0$ (since $\nabla \LL(\w_t)$ and $\xib_t$ and $\zetab_t$ are independent),
we obtain: 
\begin{align}
    \sum_{t=1}^{\Te}(\eta_t-\frac{L}{2}\eta_t^2) \mathbb{E}_{\Psi_N}[\|\nabla \LL(\w_t)\|^2 ]
    &\leq \LL(\w_0)\!-\LL^*\!
    -\!\sum_{t=1}^{\Te} (\eta_t-L\eta_t^2)\mathbb{E}_{\Psi_N}[\left< \nabla \LL ( \w_t ), \xib_t \right >] %
    +\sum_{t=1}^{\Te}\frac{L}{2}\eta_t^2(\frac{\sigma^2}{r}+%
    \mathbb{E}_{\Psi_N}[\|\xib_t\|^2]).
\end{align}
Next, we analyze convergence under two cases: (1) where $\E[\|\xib_t\|]\leq \epsilon \|\nabla \LL(\w_t)\|$, and (2) where $\E[\|\xib_t\|]\leq \epsilon$.
\paragraph{Case 1. Assuming 
$\E[\|\xib_t\|]\leq \epsilon \|\nabla \LL(\w_t)\|$ for $0\leq \epsilon < 1$.} With $1/L\leq\eta_t<2/L$, we have $\eta_t-L\eta^2_t\leq 0$. Hence, %
\begin{align}
    \sum_{t=1}^{\Te}(\eta_t\!-\!\frac{L}{2}\eta_t^2) \mathbb{E}_{\Psi_N}[\|\nabla \LL(\w_t)\|^2 ]
     &\leq \!\LL(\w_0)\!-\!\LL^*\!\!
    -\!\!\sum_{t=1}^{\Te} (\eta_t-L\eta_t^2)\epsilon \mathbb{E}_{\Psi}[\|\nabla \LL(\w_t)\|^2] %
    +\!\!\sum_{t=1}^{\Te}\frac{L}{2}\eta_t^2(\frac{\sigma^2}{r}\!+\!\epsilon^2 \mathbb{E}_{\Psi_N}[\|\nabla \LL(\w_t)\|^2]).\label{eq:norm}
\end{align}
Hence,
\begin{align}
    \sum_{t=1}^{\Te} \left(\eta_t-\frac{L}{2}\eta_t^2 + \epsilon (\eta_t-L\eta_t^2) -\frac{L}{2} \eta_t^2\epsilon^2 \right) \mathbb{E}_{\Psi_N}[\|\nabla \LL(\w_t)\|^2 ]
    &\leq \LL(\w_0)-\LL^* + \frac{L\sigma^2}{2r}\sum_{t=1}^{\Te} \eta_t^2,\\
    \sum_{t=1}^{\Te} \left((1+\epsilon)\eta_t-\frac{L}{2}(1 + \epsilon)^2 \eta_t^2\right) \E_{\Psi_N}[\|\nabla \LL(\w_t)\|^2 ]
    &\leq \LL(\w_0)-\LL^* + \frac{L\sigma^2}{2r}\sum_{t=1}^{\Te} \eta_t^2,
\end{align}
and we get: 
\begin{align}
    \mathbb{E}_{\Psi_N}[\|\nabla \LL(\w_t)\|^2 ]\leq\frac{1}{\sum_{t=1}^{\Te} (1+\epsilon)\eta_t- \frac{L}{2}(1 + \epsilon)^2 \eta_t^2} \left[{\LL(\w_0)-\LL^*}+ \frac{\sigma^2L}{2r}\sum_{t=1}^{\Te} \eta_t^2\right]
\end{align}
If $\eta_t< %
(1+\epsilon)/(\frac{L}{2}(1+\epsilon)^2)=2/L(1+\epsilon)$, then $(1+\epsilon)- \frac{L}{2}(1 + \epsilon)^2 \eta_t> 0$ and we have: 
\begin{align}
    \mathbb{E}_{\Psi_N}[\|\nabla \LL(\w_t)\|^2 ]
    &\leq\frac{1}{\sum_{t=1}^{\Te} \eta_t} \left[{2(\LL(\w_0)-\LL^*)}+ \frac{\sigma^2L}{r}\sum_{t=1}^{\Te} \eta_t^2\right].
\end{align}
For a random iterate $R$ of a run of the algorithm that is selected with probability $(2(1+\epsilon)\eta_t-L(1+\epsilon)^2\eta_t^2)/\sum_{t=1}^N(2(1+\epsilon)\eta_t-L(1+\epsilon)^2\eta_t^2)$, we have that
\begin{align}
    \E[\|\nabla\LL(\w_R)\|^2]=\E_{R,\Psi_N}[\|\nabla\LL(\w_R)\|^2]=\frac{\sum_{t=1}^N(2(1+\epsilon)\eta_t-L(1+\epsilon)^2\eta_t^2)\E_{\Psi_N}[\|\nabla\LL(\w_t)\|^2]}{\sum_{t=1}^N(2(1+\epsilon)\eta_t-L(1+\epsilon)^2\eta_t^2)}=\E_{\Psi_N}[\|\nabla\LL(\w_t)\|^2]
\end{align}
Hence, for $\eta_t=\eta$ we get:
\begin{align}
    \mathbb{E}[\|\nabla \LL(\w_R)\|^2 ]
    &\leq\frac{1}{\Te \eta} \left[{2(\LL(\w_0)-\LL^*)}+ \frac{\sigma^2L}{r}\Te \eta^2\right]. \label{eq:grad_bound}
\end{align}

For $\eta=\min\{\frac{1}{L},\frac{\tilde{D}\sqrt{r}}{\sigma \sqrt{N}}\}$, and $\tilde{D}>0$, we get 
\begin{align}
    \mathbb{E}[\|\nabla \LL(\w_R)\|^2 ]
    &\leq\frac{1}{\Te \eta} \left[{2(\LL(\w_0)-\LL^*)}\right]+ \frac{\sigma^2L}{r} \eta\\
    &\leq \frac{2(\LL(\w_0)-\LL^*)}{N}\max\{L,\frac{\sigma \sqrt{N}}{\tilde{D}\sqrt{r}}\}+\frac{\sigma^2L}{r}\frac{\tilde{D}\sqrt{r}}{\sigma \sqrt{N}}\\
    &\leq \frac{2L(\LL(\w_0)-\LL^*)}{N}+\left(L\tilde{D}+\frac{2(\LL(\w_0)-\LL^*)}{\tilde{D}}\right)\frac{\sigma}{ \sqrt{rN}}
\end{align}
Replacing the optimal value $\tilde{D}=\sqrt{2(\LL(\w_1)-\LL^*)/L}$, we get
\begin{align}
    \mathbb{E}[\|\nabla \LL(\w_R)\|^2 ]
    &\leq \mathcal{B}_N:=\frac{2L(\LL(\w_0)-\LL^*)}{N}+\frac{2\sigma\sqrt{2L(\LL(\w_0)-\LL^*)}}{ \sqrt{rN}}
\end{align}
Hence, training with \alg exhibits an $\OO(1/\sqrt{rN})$ rate of convergence, compared to $\OO(1/\sqrt{mN})$ for mini-batch SGD with mini-batch size $m<r$.

To derive large-deviation properties for a single run of this method, we are interested in the number of iterations required to find a point satisfying $\mathbb{P}[\|\nabla\LL(\w_R)\|^2\leq \nu^2]\geq 1-\frac{1}{\lambda}$. We use Markov's inequality to calculate the probability $\mathbb{P}[\|\nabla\LL(\w_R)\|^2\geq \lambda  \mathcal{B}_N]\leq\frac{1}{\lambda}$. We get that %
with probability at least $1-\lambda$, at least one iteration of a single run of the algorithm visits a $\nu$-stationary point in the following number of iterations:
\begin{align}
    \tilde{\mathcal{O}}\left(\frac{L(\LL(\w_0)-\LL^*)}{{\nu}^2}(1+\frac{\sigma^2}{r{\nu}^2})\right).
\end{align}
As long as $r\leq {\sigma^2}/{\nu^2}$ increasing the size of the random subsets $r$ used by \alg will reduce the number of iterations linearly, while not increasing the total number of stochastic gradient queries.

\paragraph{Case 2. Assuming $\E[\|\xib_t\|]\leq \epsilon<\nu^2$.}
For $1/L\leq\eta_t<2/L$ we have $\eta_t-L\eta^2_t<0$. Hence:
\begin{align}
    \sum_{t=1}^{\Te}(\eta_t-\frac{L}{2}\eta_t^2) \mathbb{E}_{\Psi}[\|\nabla \LL(\w_t)\|^2 ]
    &\leq \LL(\w_0)-\LL^*
    -\sum_{t=1}^{\Te} (\eta_t-L\eta_t^2)\mathbb{E}_{\Psi}[\left< \nabla \LL ( \w_t ), \xib_t \right >] %
    +\sum_{t=1}^{\Te}\frac{L}{2}\eta_t^2(\frac{\sigma^2}{r}+\epsilon^2 ),\\
    &\leq \LL(\w_0)-\LL^*
    -\sum_{t=1}^{\Te} (\eta_t-L\eta_t^2)\mathbb{E}_{\Psi}[\left|\left< \nabla \LL ( \w_t ), \xib_t \right >\right|] %
    +\sum_{t=1}^{\Te}\frac{L}{2}\eta_t^2(\frac{\sigma^2}{r}+\epsilon^2 ),\\
     &\leq \LL(\w_0)-\LL^*
    -\sum_{t=1}^{\Te} (\eta_t-L\eta_t^2)\epsilon \mathbb{E}_{\Psi}[\|\nabla \LL(\w_t)\|] %
    +\sum_{t=1}^{\Te}\frac{L}{2}\eta_t^2(\frac{\sigma^2}{r}+\epsilon^2 )\\
    &\leq \LL(\w_0)-\LL^*
    -\sum_{t=1}^{\Te} (\eta_t-L\eta_t^2)\epsilon \nabla_{\max}
    +\sum_{t=1}^{\Te}\frac{L}{2}\eta_t^2(\frac{\sigma^2}{r}+\epsilon^2 ),
\end{align}
where $\nabla_{\max}=\max\{0,\max_{i\in V,\w_t\in\W} \|\g_{t,i}\|\}$. 
For a random iterate $R$ of a run of the algorithm that is selected with probability $(2\eta_t-L\eta_t^2)/\sum_{t=1}^N(2\eta_t-L\eta_t^2)$,  we have that
\begin{align}
    \E[\|\nabla\LL(\w_R)\|^2]=\E_{R,\Psi_N}[\|\nabla\LL(\w_R)\|^2]=\frac{\sum_{t=1}^N(2\eta_t-L\eta_t^2)\E_{\Psi_N}[\|\nabla\LL(\w_t)\|^2]}{\sum_{t=1}^N(2\eta_t-L\eta_t^2)}=\E_{\Psi_N}[\|\nabla\LL(\w_t)\|^2]
\end{align}
If $\eta=\eta_t$ and $\eta\leq1/L+1/2L\nabla_{\max}$, we have $-2(1-L\eta)\leq1/\nabla_{\max}$ and we get
\begin{align}
     \mathbb{E}[\|\nabla \LL(\w_{R})\|^2 ]
    &\leq\frac{1}{\Te \eta (1-\frac{L}{2}\eta)} \left[{\LL(\w_0)-\LL^*}-\Te(\eta-L\eta^2)\epsilon\nabla_{\max}+ (\frac{\sigma^2}{r}+\epsilon^2)\frac{L}{2}\Te \eta^2\right]\\
    &\leq\frac{2}{\Te \eta} \left[{\LL(\w_0)-\LL^*}-\Te(\eta-L\eta^2)\epsilon\nabla_{\max}+ (\frac{\sigma^2}{r}+\epsilon^2)\frac{L}{2}\Te \eta^2\right]\\
    &=\frac{2}{\Te \eta} \left[{\LL(\w_0)-\LL^*}\right]-2(1-L\eta)\epsilon\nabla_{\max}+ (\frac{\sigma^2}{r}+\epsilon^2)L \eta\\
    &\leq\frac{2}{\Te \eta} \left[{\LL(\w_0)-\LL^*}\right]+\epsilon+ (\frac{\sigma^2}{r}+\epsilon^2)L \eta \label{eq:eps}
\end{align}
For $\eta=\min\{\frac{1}{L},\frac{\tilde{D}}{\sqrt{N(\sigma^2/r+\epsilon^2)}}\}$, and $\tilde{D}>0$, we get 
\begin{align}
    \mathbb{E}[\|\nabla \LL(\w_R)\|^2 ]
    &\leq\frac{1}{\Te \eta} \left[{2(\LL(\w_0)-\LL^*)}\right]+  (\frac{\sigma^2}{r}+\epsilon^2)L \eta+\epsilon\\
    &\leq \frac{2(\LL(\w_0)-\LL^*)}{N}\max\{L,\frac{\sqrt{N(\sigma^2/r+\epsilon^2)}}{\tilde{D}}\}+(\frac{\sigma^2}{r}+\epsilon^2) \frac{L\tilde{D}}{\sqrt{N(\sigma^2/r+\epsilon^2)}}+\epsilon\\
    &\leq \frac{2L(\LL(\w_0)-\LL^*)}{N}+\left(L\tilde{D}+\frac{2(\LL(\w_0)-\LL^*)}{\tilde{D}}\right)\frac{\sqrt{\sigma^2/r+\epsilon^2)}}{ \sqrt{N}}+\epsilon
\end{align}
For the optimal value of $\tilde{D}=\sqrt{2(\LL(\w_1)-\LL^*)/L}$, we get
\begin{align}
    \mathbb{E}[\|\nabla \LL(\w_R)\|^2 ]
    &\leq \mathcal{B}_N:=\frac{2L(\LL(\w_0)-\LL^*)}{N}+\frac{2\sqrt{\sigma^2+r\epsilon^2}\sqrt{2L(\LL(\w_0)-\LL^*)}}{ \sqrt{rN}}+\epsilon
\end{align}
Hence, the number of iterations becomes: 
\begin{align}
    \tilde{\mathcal{O}}\left(\frac{L(\LL(\w_0)-\LL^*)}{\nu^2-\epsilon}(1+\frac{\sigma^2+r\epsilon^2}{r(\nu^2-\epsilon)})\right)
\end{align}
Hence, more number of iterations is required. Besides, if $\epsilon\geq \nu^2$, convergence is not guaranteed.

\paragraph{Incorporating $\tau$.}
Assume $c_2$ is the error of the coreset in capturing the full gradient at the beginning of the neighborhood. From Eq. \eqref{eq:threshold} we know
$|\LL(\w_{t_l}+\del) - \F^l(\del)| 
    =\rho_{t_l}\LL(\w_{t_l}+\del_l)$. Using the quadratic approximation in Eq. \eqref{eq:m}, i.e., $\F^{l}(\del)=\frac{1}{2}\del^T\HH_{t_l,S_l}\del+\g_{t_l,S_l}\del+\LL(\w_{t_l})$, and noting that $\LL$ can also be modeled by a similar quadratic function for small $\tau$, we get:
\begin{align}
    \rho_{t_l}\LL(\w_{t_l}+\del_l)= |\LL(\w_{t_l}+\del_l) - \F^l(\del_l)|
    &=|\frac{1}{2}\del_l^T(\HH_{t_l,V}^{}-\HH_{t_l,S_l})\del_l+(\g_{t_l,V}-\g_{t_l,S_l}^{})\del_l|\\
    &\geq \Big|\frac{1}{2} |\del_l^T(\HH_{t_l,V}^{}-\HH_{t_l,S_l}^{})\del_l|-\|\g_{t_l,V}^{}-\g_{t_l,S_l}^{}\|\cdot\|\del_l\| \Big |\\
    &\geq \Big|\frac{1}{2} |\del_l^T(\HH_{t_l,V}^{}-\HH_{t_l,S_l}^{})\del_l|-c_2\cdot\|\del_l\| \Big |\\
    &\geq \frac{1}{2} |\del_l^T(\HH_{t_l,V}^{}-\HH_{t_l,S_l}^{})\del_l|-c_2\cdot\|\del_l\| 
\end{align}
As long as $\rho_{t_l}$ is small, we can assume that the loss can be well modeled by a quadratic using the Hessian diagonal.
Using the Hessian diagonal {for both} $\LL$ and $\F^l$, we have $|\del_l^T(\HH_{t_l,S_l}^{}-\HH_{t_l,V}^{})\del_l|=\|\del_l^T(\text{diag}(\HH_{t_l,S_l}^{})-\text{diag}(\HH_{t_l,V}))\|\cdot\|\del_l\|$. Hence,
\begin{align}
    &\frac{1}{2}|\del_l^T(\HH_{t_l,S_l}^{}-\HH_{t_l,V}^{})\del_l|=\frac{1}{2}\|\del^T(\text{diag}(\HH_{t_l,S_l}^{})-\text{diag}(\HH_{t_l,V}^{}))\|\cdot\|\del_l\|\leq \rho_{t_l}\LL(\w_{t_l}+\del_l) +c_2\|\del_l\|,
    \\
    &\|\del_l^T(\text{diag}(\HH_{t_l,S_l}^{})-\text{diag}(\HH_{t_l,V}^{}))\|\leq 2\rho_{t_l}\LL(\w_{t_l}+\del_l)/\|\del_l\| +2c_2.
    \label{eq:up_bound}
\end{align}
On the other hand, we have that $\nabla\F^l(\del)=\del^T\HH_{t_l,S_l}^{}+\g_{t_l,S_l}^{}$ and $\nabla\LL(\w_{t_l}+\del_l)=\del_l^T\HH_{t_l,V}^{}+\g_{t_l,V}^{}$. Hence, we have:
\begin{align}
    \|\nabla \LL(\w_{t_l}+\del_l) - \nabla \F^l(\del_l)\| &= \|\del_l^T(\HH_{t_l,V}^{}-\HH_{t_l,S_l}^{})+(\g_{t_l,V}^{}-\g_{t_l,S_l}^{})\|\\
    &\leq \|\del_l^T(\HH_{t_l,V}^{}-\HH_{t_l,S_l}^{})\|+\|\g_{t_l,V}^{}-\g_{t_l,S_l}^{}\|\\
    &\leq \|\del_l^T(\HH_{t_l,V}^{}-\HH_{t_l,S_l}^{})\|+c_2.\label{eq:g_upper}
\end{align}
Therefore, using Hessian diagonal and from Eq. \eqref{eq:up_bound} and \eqref{eq:g_upper} we get:
\begin{equation}
    \|\nabla \LL(\w_{t_l}+\del_l) - \nabla \F^l(\del_l)\| \leq \|\del_l^T(\text{diag}(\HH_{t_l,V}^{})-\text{diag}(\HH_{t_l,S_l}^{}))\|+c_2\leq {2\rho_{t_l} \LL(\w_{t_l}+\del_l)}/{\|\del_l\|}+3c_2 %
    \label{eq:g_error}
\end{equation}
Eq. \eqref{eq:g_error} shows that for a fixed $\rho_{t_l}$ and loss, if the convex approximation $\F^l$ is valid in a larger neighborhood $\del_l$, then the error of the Hessian diagonal at the beginning of the neighborhood was smaller and hence the gradient error at the end of the neighborhood %
is smaller.

From Eq. \eqref{eq:g_error} we know that $\|\nabla \LL(\w_{t_l}+\del_l) - \nabla \F^l(\del_l)\|\leq 2\rho_{t_l} \LL(\w_{t_l}+\del_l)/\|\del_l\|+3c_2$. Let %
$c_1$ be the desired upper-bound on the gradient error at $\w_{t_l}+\del_l$. Hence, we wish
\begin{equation}
\|\nabla \LL(\w_{t_l}+\del_l) - \nabla \F^l(\del_l)\|\leq 2\rho_{t_l} \LL(\w_{t_l}+\del_l)/\|\del_l\|+3 c_2\leq c_{1}. 
\end{equation}
Hence,  for $c_2\leq c_1/3$ we get:
\begin{equation}
\rho_{t_l} \leq \frac{(c_1-3c_2)\|\del_l\|}{2\LL(\w_{t_l}+\del_l)}, \quad\quad  \tau\leq\min_{t_l} \rho_{t_l}.
\end{equation} 
For $\|\nabla \LL(\w_{t_l}+\del_l) - \nabla \F^l(\del_l)\|\leq  c_1=c \|\nabla \LL(\w_{t_l}\!+\!\del_l)\|$, and with $c_2\leq c\|\nabla \LL(\w_{t_l}\!+\!\del_l)\|/3$, we have:
\begin{equation}
\rho_{t_l} \leq \frac{(c\|\nabla\LL(\w_{t_l}+\del_l)\|-3c_2)\|\del_l\|}{2\LL(\w_{t_l}+\del_l)}, %
\end{equation} 
For $c=1$ and $c_2=\epsilon\|\nabla \LL(\w_t)\|$, we get Case 1 in the Theorem.
We see that when gradient norm is smaller, we should have a smaller error $c_2$ in capturing the random subset gradients.

\subsection{Experimental details}\label{appendix:details}
\begin{table}[h]
\caption{Experiment setups. }
\label{tab:dataset}
\vspace{-3mm}
\begin{center}
\begin{small}
\begin{sc}
\begin{tabular}{lrrrr|r}
\toprule
Dataset & Classes & Train & Network & Parameters & Ful acc\\
\midrule
CIFAR-10 & 10 & 50k & ResNet-20 & 0.27M &  $92.1_{\pm 0.1}$\\
CIFAR-100 & 100 & 50k & ResNet-18 & 11M & $75.6_{\pm 0.3}$ \\
TinyImageNet & 200 & 100k & ResNet-50 & 23M &$66.9_{\pm 0.1}$ \\
\midrule
SNLI & 3 & 570k & RoBERTa & 123M & $92.9_{\pm 0.2}$ \\
\bottomrule
\end{tabular}
\end{sc}
\end{small}
\end{center}
\vskip -0.1in
\end{table}

\begin{table}[t]
\caption{Relative error (\%) with 20\% of the full training budget (backprop) can reach a very close accuracy to that of full training (with only 2-3\% difference) on all datasets, namely, CIFAR-10, CIFAR-100, and TinyImageNet.  %
}
\label{tab:20-acc}
\vspace{-3mm}
\begin{center}
\begin{small}
\begin{sc}
\begin{tabular}{lccc}
\toprule
 & \alg & Random & SGD\textdagger \\
 \midrule
 CIFAR-10 ~-~ ResNet-20 & 2.32 & 2.87 & 16.47 \\
 CIFAR-100 ~-~ ResNet-18 & 3.37 & 3.66 & 32.68 \\
 TinyImageNet ~-~ ResNet-50 & 3.05 & 3.51 & 47.43 \\
\bottomrule
\end{tabular}

\end{sc}
\end{small}
\end{center}
\vskip -0.1in
\end{table}

\textbf{Tuning Hyperparameters.}
We tuned the hyperparameters $\tau \in \{0.1, 0.05, 0.01, 0.005, 0.001\}$, $h \in \{1,2,4,8,10\}$ and
used $\tau=0.05, 0.01, 0.005, 0.05$, $~h=1,10,1,4$ on CIFAR-10, CIFAR-100, TinyImagenet, and SNLI, respectively, as listed in \cref{tab:hyperparam}. To determine $\tau$, we calculated the average loss approximation error divided by the training loss, i.e. $\rho_{t_l}$ in Eq. \eqref{eq:threshold}, after some coresets updates during training. 
Across all datasets, we found that $\alpha=0.1$ yielded satisfactory results.

\begin{figure*}[t]
     \centering
     \begin{subfigure}[t]{0.4\textwidth}
         \centering
         \includegraphics[width=\textwidth]{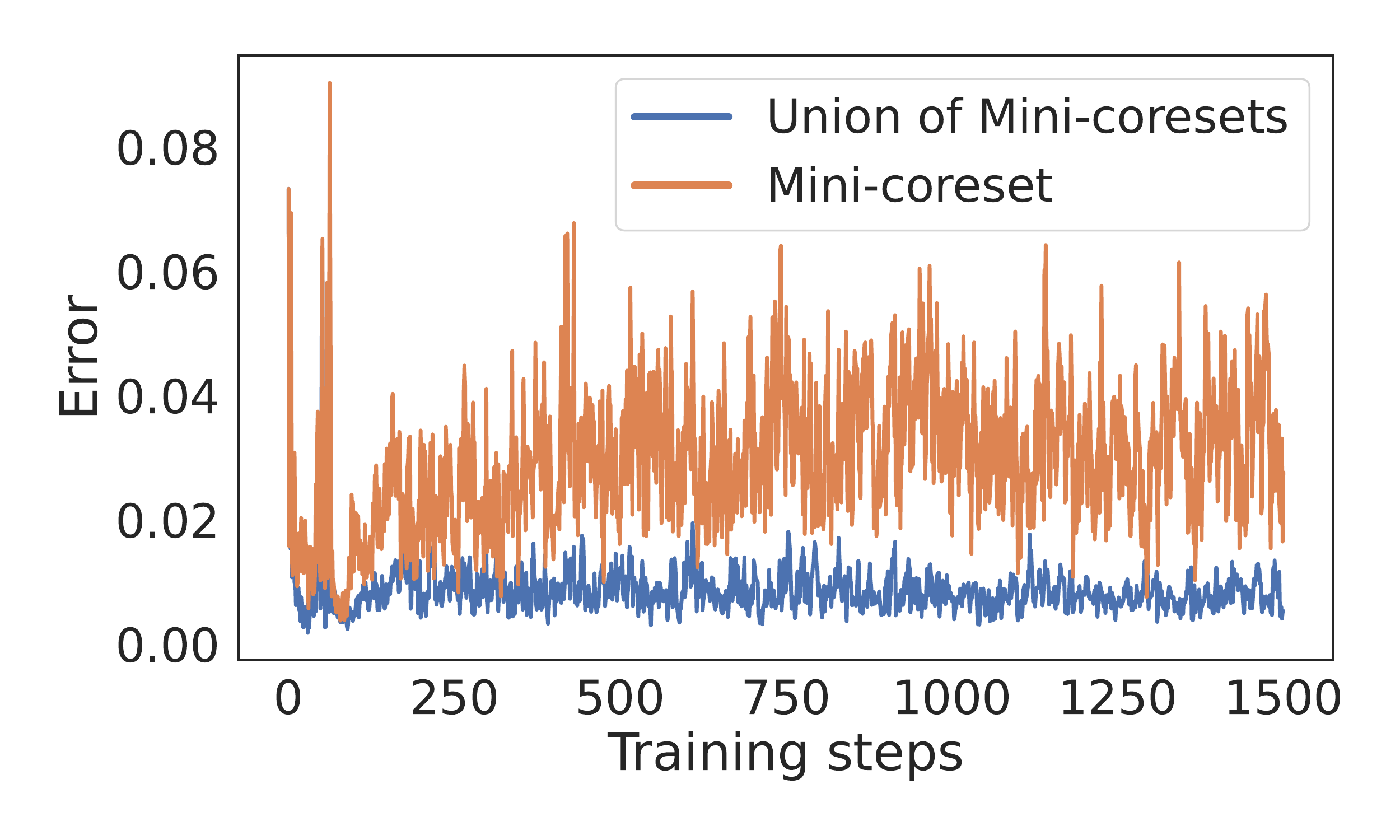}\vspace{-3mm}
         \caption{}
         \label{fig:union}
     \end{subfigure}
     \begin{subfigure}[t]{0.4\textwidth}
         \centering
         \includegraphics[ width=\textwidth]{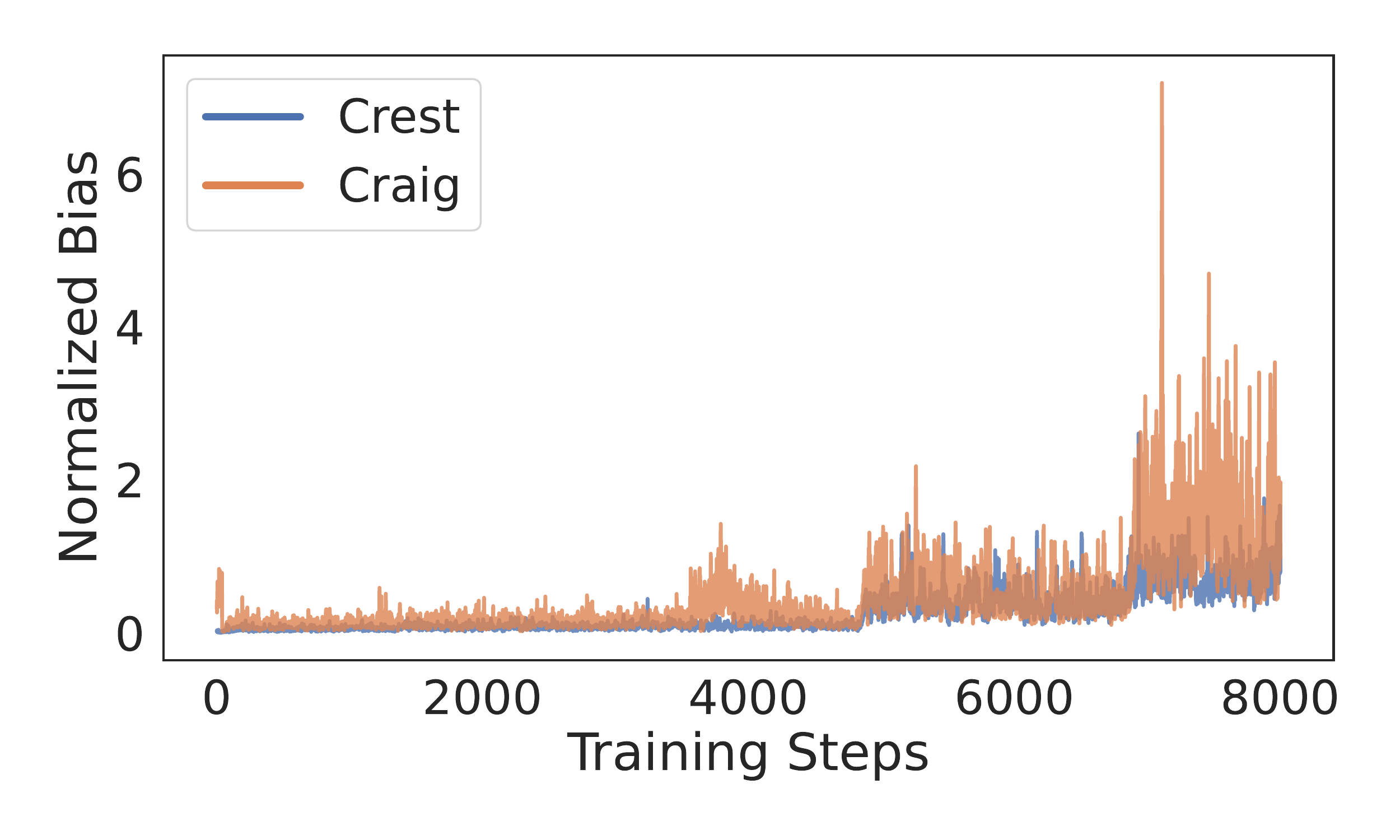}\vspace{-3mm}
         \caption{}
         \label{fig:bias_normed}
     \end{subfigure}
        \caption{Training ResNet-20 on CIFAR-10. (a) Union of mini-batch coresets has a smaller error in capturing the full gradient, compared to the bias of the individual mini-batch coresets. (b) Normalized bias of coresets by the full gradient norm, i.e., $\epsilon=\E[\|\xib_{t_l}\|]/ \|\nabla \LL(\w_{t_l})\|$ in Theorem \ref{thm}. \alg coresets have a consistently small $\epsilon<1$. As the gradient norm becomes smaller closer to the stationary points, small $\epsilon$ implies that the bias of the \alg mini-batch coresets $\E[\|\xib_{t_l}\|]$ diminishes closer to the stationary points. 
        Hence, convergence of \alg can be guaranteed (Case 1 in Theorem \ref{thm}). On the other hand, $\epsilon$ can be large for \craig coresets. Hence, convergence is not guaranteed (Case 2 in Theorem \ref{thm}).  }
        \label{fig:}
        \vspace{-3mm}
\end{figure*}

\textbf{Convergence of \alg vs \craig.} \cref{fig:bias_normed} shows training ResNet-20 on CIFAR-10 with \alg vs \craig. We see that the normalized bias of \alg mini-batch coresets over full gradient norm, i.e., $\epsilon=\E[\|\xib_{t_l}\|]/ \|\nabla \LL(\w_{t_l})\|$ is consistently small ($<1$) during the training. As the gradient norm becomes smaller closer to a stationary point, small $\epsilon$ implies that the bias of the \alg mini-batch coresets $\E[\|\xib_{t_l}\|]$ diminishes as we get closer to a stationary point. Hence, convergence of \alg can be guaranteed (Case 1 in Theorem \ref{thm}). On the other hand, the normalized error for \craig coresets can be large during the training. Hence, convergence is not guaranteed (Case 2 in Theorem \ref{thm}).

\begin{figure}[t]
\begin{minipage}[c][][c]{0.49\textwidth}
\caption{Relative error (\%) with 10\% training budget. Training on \alg mini-batch coresets of size $m=128$ selected from random subsets $V_p$ of size $r=500$ has a smaller relative error than training on random mini-batches of size $m=128$. In particular, relative error of \alg with $m=128$ is close to that of training on random mini-batches of size $m=500$. %
}
\label{tab:acc-m}
\begin{center}
\begin{small}
\begin{sc}
\begin{tabular}{lc}
\toprule
Method & Relative Error\\
\midrule
\alg m=128 &  5.2 \\
Random m=128 & 7.1  \\
Random m=512 & 4.0 \\
\bottomrule
\end{tabular}
\end{sc}
\end{small}
\end{center}
\end{minipage}
\begin{minipage}[c][][c]{0.5\textwidth}
\centering
\includegraphics[width=0.8\textwidth]{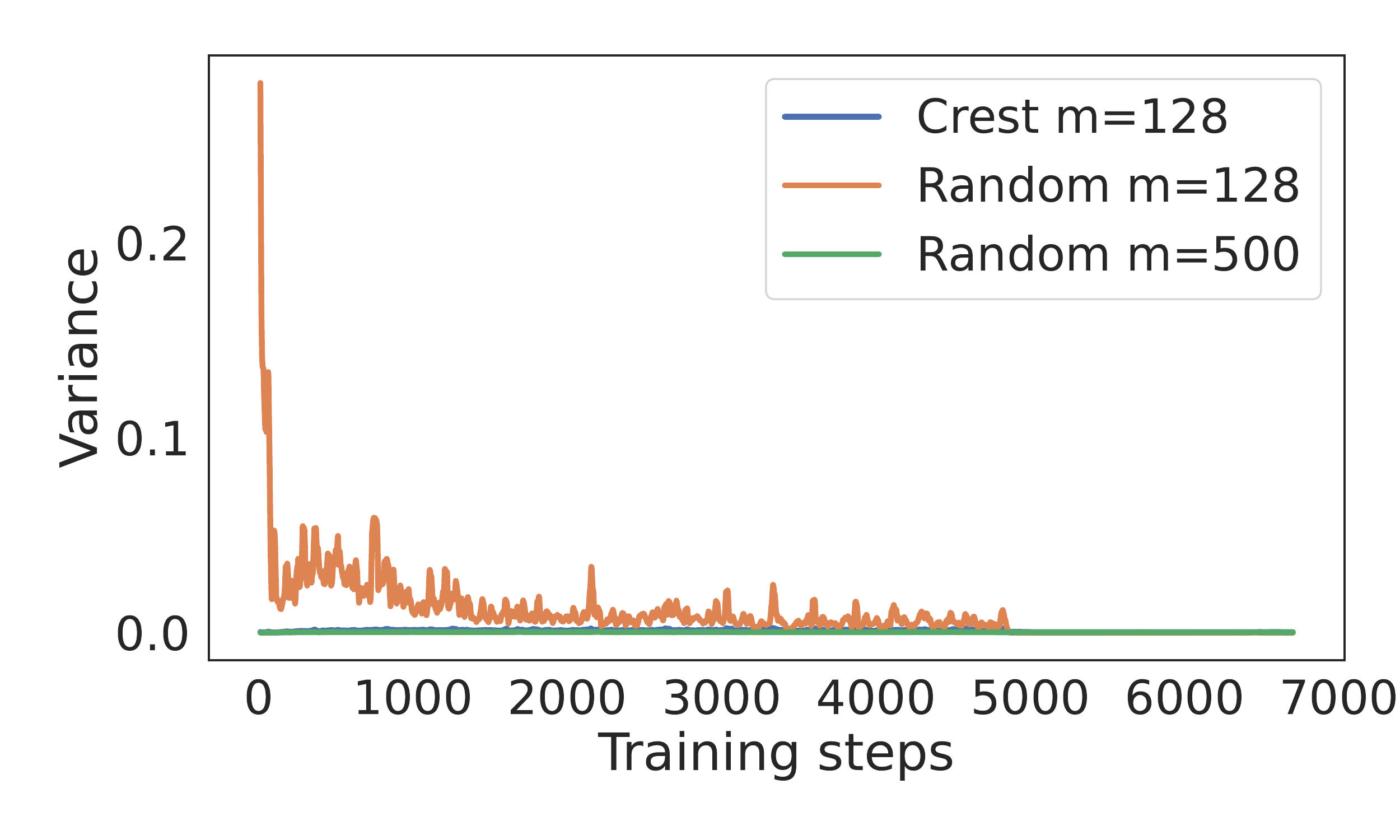}\vspace{-3mm}
\caption{Variance of gradients of \alg mini-batches of size $m=128$ selected from random subsets $V_p$ of size $r=500$ is very closer to the variance of $V_r$. In contrast, random subsets of size $m=128$ have a considerably larger variance.}
\label{fig:3_full}
\end{minipage}
\end{figure}

\textbf{\alg has a Similar Performance to Training with Large Mini-batches.} 
\cref{fig:3_full} shows the variance of gradient of \alg mini-batch coresets of size $m=128$ selected from random subsets $V_p$ of size $r=500$. We see that the variance of \alg mini-batch coresets is very close to the variance of $V_r$. In contrast, random subsets of size $m=128$ have a considerably larger variance.
\cref{tab:acc-m} further compares the relative error of \alg with mini-batch coresets of size $m=128$ selected from random subsets of size $r=500$. We see that training on \alg mini-batch coresets has a smaller relative error than training on random mini-batches of size $m=128$. In particular, relative error of \alg with $m=128$ is close to that of training on random mini-batches of size $m=500$. This is due to the smaller gradient variance of \alg mini-batch coresets, as is shown in \cref{fig:3_full}.

\begin{figure*}[t]
     \centering
     \begin{subfigure}[t]{0.4\textwidth}
         \centering
         \includegraphics[width=\textwidth]{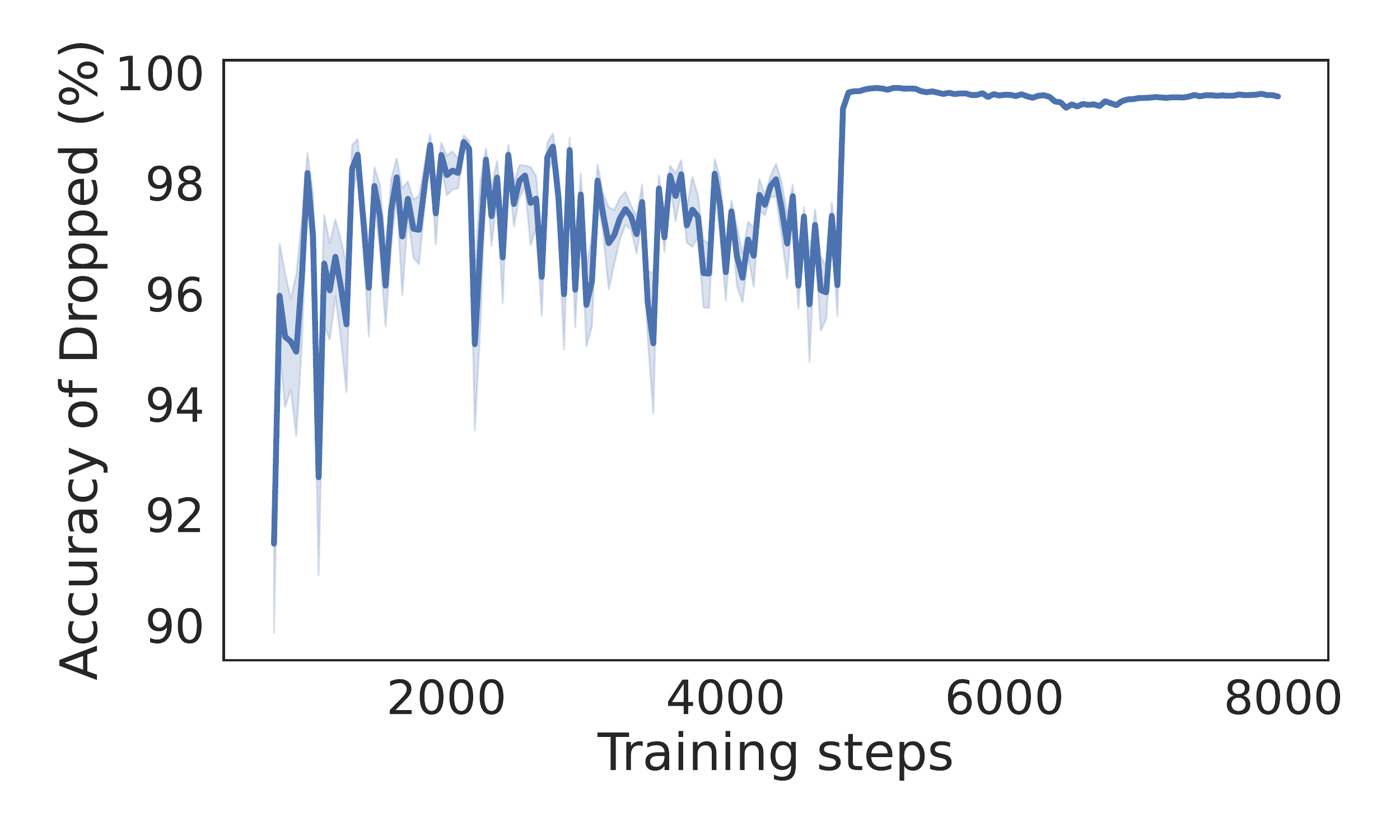}\vspace{-3mm}
         \caption{}
         \label{fig:drop_acc}
     \end{subfigure}
     \begin{subfigure}[t]{0.4\textwidth}
         \centering
         \includegraphics[width=\textwidth]{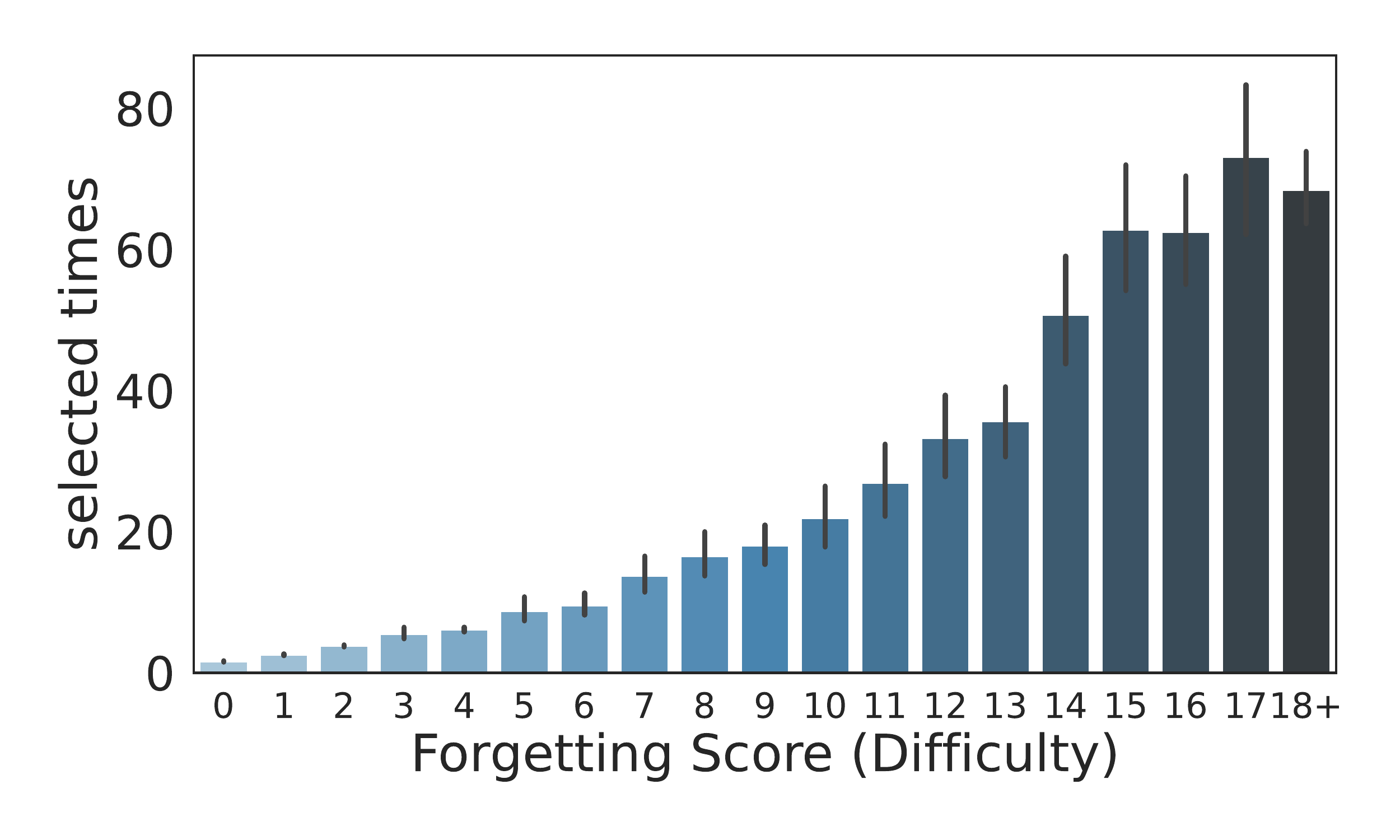}\vspace{-3mm}
         \caption{}
         \label{fig:forget}
     \end{subfigure}
        \caption{Training ResNet-20 on CIFAR-10 with \alg. (a) Dropped examples are learned later in training, by training on \alg subsets. (b) Distribution of forgetting scores for the examples selected by \alg during the training. The distribution is long-tailed, confirming that not all examples contribute equally to training.}
        \label{fig:}
        \vspace{-3mm}
\end{figure*}

\textbf{Effect of Dropping the Learned Examples.} By tracking the prediction accuracy of the dropped training examples (\cref{fig:drop_acc}), we found that even though some of the dropped examples could be forgotten after being dropped (the accuracy of the dropped examples is ~92\% earlier in training), they can be learned again when training on the coresets selected from the remaining training examples (the accuracy of dropped examples always increases to above 99\% even though we never train on them again). This confirms that dropping the learned examples does not harm the performance. 

\textbf{\alg with Larger Training Budget.}
In general, coreset methods are most beneficial under a limited training budget.
\cref{tab:20-acc} compares the relative error of training ResNet-18 on CIFAR-10, ResNet-20 on CIFAR-100 and ResNet-50 on TinyImagenet with \alg vs. Random, under 20\% training budget.
Note that under the standard learning rate schedule used for training on the above datasets for 200 epochs, there is a large gap up to 44.38\% between SGD\textdagger\ (i.e., training for $20\% \times 200=40$ epochs on full data with mini-batch SGD) and \alg. But, the gap reduces when learning rate drops at 60\% and 85\% of training (Random vs. \alg).

\begin{table}[h]
\caption{Hyperparameters used for different datasets. \looseness=-1
}
\label{tab:hyperparam}
\vspace{-3mm}
\begin{center}
\begin{small}
\begin{sc}
\begin{tabular}{lcc}
\toprule
Dataset & $\tau$ & $h$\\
\midrule
CIFAR-10 & 0.05 & 1 \\
CIFAR-100 &  0.01 & 10 \\
TinyImagenet & 0.005 & 1 \\
SNLI & 0.05 & 4\\
\bottomrule
\end{tabular}
\vspace{-3mm}
\end{sc}
\end{small}
\end{center}
\end{table}

\end{document}